\documentclass[runningheads,a4paper]{llncs}

\usepackage{amssymb}
\usepackage{amsmath,bm}
\usepackage{pifont}
\usepackage{graphicx}
\usepackage{url}
\usepackage{xspace}
\usepackage{hyperref}
\usepackage{xcolor}
\usepackage{booktabs} 
\usepackage{makecell}
\usepackage{paralist, tabularx}
\usepackage{algorithm}
\usepackage{algorithmic}
\usepackage{caption}
\usepackage{subcaption}
\usepackage[T1]{fontenc}
\usepackage{multirow,multicol}
\usepackage{wrapfig}
\usepackage{appendix}

\newcommand{\method}{{\sc SPI}\xspace}
\newcommand{\kt}{{\sc FT}\xspace}
\newcommand{\spil}{{\sc SPI-lite}\xspace}

\newcommand{\oc}{{\sc OC-SVM+}\xspace}
\newcommand{\ocr}{{\sc OC-SVM}\xspace}
\newcommand{\ifr}{{\sc iF}\xspace}

\definecolor{OliveGreen}{rgb}{0,0.6,0}
\newcommand{\tick}{\textcolor{OliveGreen}{\ding{51}}}
\makeatletter
\newcommand\footnoteref[1]{\protected@xdef\@thefnmark{\ref{#1}}\@footnotemark}
\makeatother

\newcommand{\cbit}{\begin{compactitem}}
	\newcommand{\ceit}{\end{compactitem}}
\newcommand{\cben}{\begin{compactenum}}
	\newcommand{\ceen}{\end{compactenum}}

\newcommand{\bit}{\begin{itemize}}
	\newcommand{\eit}{\end{itemize}}
\newcommand{\ben}{\begin{enumerate}}
	\newcommand{\een}{\end{enumerate}}
\newcommand{\beq}{\begin{equation}}
\newcommand{\eeq}{\end{equation}}

\newcommand{\bz}{\boldsymbol{z}}
\newcommand{\bzs}{\boldsymbol{z^{*}}}
\newcommand{\bx}{\boldsymbol{x}}
\newcommand{\bxs}{\boldsymbol{x^{*}}}
\newcommand{\bphi}{\boldsymbol{\phi}}
\newcommand{\bgamma}{\boldsymbol{\gamma}}

\newcommand{\bbeta}{\boldsymbol{\beta}}

\newcommand{\hide}[1]{}

\newsavebox\CBox
\def\textBF#1{\sbox\CBox{#1}\resizebox{\wd\CBox}{\ht\CBox}{\textbf{#1}}}

\def\iforest {\textsc{iF}}
\def\iforestpi{\textsc{iF}\textsuperscript{*}}
\def\ocsvm {\textsc{OC+}}
\def\spi{\textsc{SPI}}
\def\spiscores {\textsc{SPI-lite}}
\def\ft {\textsc{FT}}

\hypersetup{
	colorlinks,
	citecolor={black},
	urlcolor={blue}
}

\title{Incorporating Privileged Information to \\ Unsupervised Anomaly Detection}

\author{Shubhranshu Shekhar \quad Leman Akoglu}
\institute{Carnegie Mellon University \\
 Heinz College of Information Systems and Public Policy \\
   \email{\{shubhras, lakoglu\}@andrew.cmu.edu}}

\date{}

\begin{document}
\maketitle

\begin{abstract} 
We introduce a new unsupervised anomaly detection ensemble called \method which can harness \textit{privileged} information---data available only for training examples but not for (future) test examples.
Our ideas build on the Learning Using Privileged Information (LUPI) paradigm pioneered by Vapnik et al. \cite{journals/nn/VapnikV09,conf/slds/VapnikI15}, which we extend to unsupervised learning and in particular to anomaly detection.
\method (for Spotting anomalies with Privileged Information) constructs a number of frames/fragments of knowledge 
(i.e., density estimates) in the privileged space and \textit{transfers} them to the anomaly scoring space through ``imitation'' functions that use only the partial information available for test examples.
\sloppy{
	Our generalization of the LUPI paradigm to unsupervised anomaly detection shepherds the field in several key directions, including
	($i$) 
	\textit{domain-knowledge-augmented detection} using expert annotations as PI,
	($ii$) 
	\textit{fast detection} using computationally-demanding data as PI,
	and ($iii$) 
	\textit{early detection} using ``historical future'' data as PI.
}
Through extensive experiments on simulated and real datasets, we show that augmenting privileged information to anomaly detection significantly improves detection performance.
We also demonstrate the promise of \method under all three settings ($i$--$iii$); with PI capturing expert knowledge, computationally-expensive features, and future data on three real world detection tasks.
\end{abstract}

\section{Introduction}
\label{sec:intro}
\sloppy{
	Outlier detection in point-cloud data has been studied extensively \cite{Aggarwal13}.
	In this work we consider a unique setting with a much sparser literature:
	the problem of augmenting privileged information into unsupervised anomaly detection.
	Simply put, privileged information (PI) is \textit{additional} data/knowledge/information that is available only at the learning/model building phase for (subset of) training examples, which however is unavailable for (future) test examples. 
	
	\textbf{The LUPI framework.$\;$} Learning Using Privileged Information (LUPI) has been pioneered by Vapnik et al. first in the context of SVMs \cite{journals/nn/VapnikV09,conf/slds/VapnikI15} (PI-incorporated SVM is named SVM+), later generalized to neural networks \cite{journals/amai/VapnikI17}. The setup involves an Intelligent (or non-trivial) Teacher at learning phase, who provides the Student with \textit{privileged} information (like explanations, metaphors, etc.), denoted $\bxs_i$, about each training example $\bx_i$, $i=1\ldots n$. 
	The key point in this paradigm is that \textit{privileged information is not available at the test phase} (when Student operates without guidance of Teacher).
	Therefore, the goal  is to build models (in our case, detectors) that can leverage/incorporate such additional information but \textit{yet, not depend on the availability of PI at test time}.}

\textit{Example:$\;$} The additional information $\bx^*_i$'s belong to space $X^*$ which is, generally speaking, different from space $X$. In other words, the feature spaces of vectors $\bxs_i$'s and $\bx_i$'s do not overlap.
As an example, consider the task of identifying cancerous biopsy images. Here the images are in pixel space $X$. Suppose that there is an Intelligent Teacher that can recognize patterns in such images relevant to cancer. Looking at a biopsy image, Teacher can provide 
a description like ``Aggressive proliferation of A-cells into B-cells'' or ``Absence of any dynamic''.
Note that such descriptions are in a specialized language space $X^*$, different from pixel space $X$.
Further, they would be available only for a set of examples and not when the model 
is to operate autonomously in the future.

\textit{LUPI's advantages:$\;$} LUPI has been shown to ($i$) improve rate of convergence for learning, i.e., require asymptotically fewer examples to learn \cite{journals/nn/VapnikV09}, as well as ($ii$) improve accuracy, when one can learn a model in space $X^*$ that is not much worse than the best model in space $X$ (i.e., PI is intelligent/non-trivial) \cite{journals/amai/VapnikI17}. Motivated by these advantages, LUPI has been applied to a number of problems from 
action recognition \cite{journals/ijcv/NiuLX16} to risk modeling \cite{journals/eswa/RibeiroSCVN12} (expanded in \S\ref{sec:relatedwork}).
However, the focus of all such work has mainly been on \textit{supervised} learning. 

\textbf{LUPI for anomaly detection.$\;$} The only (perhaps straightforward) extension of LUPI to unsupervised anomaly detection has been introduced recently, generalizing SVM+ to the One-Class SVM (namely \oc) \cite{conf/icdm/BurnaevS16} for malware and bot detection. The issue is that \ocr is not a reliable  detector---experiments on numerous benchmark datasets with ground truth by Emmott et al. that compared popular anomaly detection algorithms find that \ocr ranks at the bottom (Table 1, pg. 4 \cite{Emmott2013}; also see 
our results in \S \ref{sec:experiments}). We note that the top performer in \cite{Emmott2013} is the Isolation Forest (iForest) algorithm \cite{conf/icdm/LiuTZ08}, an ensemble of randomized trees.

\textit{Our contributions:$\;$} 
Motivated by LUPI's potential value to learning and the scarcity in the literature of its generalization to anomaly detection, we propose a new technique called \method (pronounced `spy'), for Spotting anomalies with Privileged Information. Our work \textit{bridges the gap (for the first time) between LUPI and unsupervised ensemble based anomaly detection} that is considered state-of-the-art \cite{Emmott2013}. 
We summarize our main contributions as follows.

\begin{itemize}
	\item {\bf Study of LUPI for anomaly detection:} 
	We analyze how LUPI can benefit anomaly detection, not only when PI is truly unavailable at test time (as in traditional setup) but also when PI is strategically and willingly avoided at test time.
	We argue that data/information that incurs overhead on resources (\$\$\$/storage/battery/etc.), timeliness, or vulnerability, if designated as PI, can enable resource-frugal, early, and preventive detection
	(expanded in \S\ref{sec:prelim}).
	
	\item \textbf{PI-incorporated detection algorithm:} We show how to 
	incorporate PI into ensemble based detectors and propose \method, which
	constructs frames/fragments of knowledge 
	(specifically, density estimates) in the privileged space ($X^*$) and \textit{transfers} them to the anomaly scoring space ($X$) through ``imitation'' functions that use only the partial information available for test examples.
	To the best of our knowledge, ours is the first attempt to leveraging PI for improving the state-of-the-art \textit{ensemble methods} for anomaly detection within an \textit{unsupervised} LUPI framework.
	Moreover, while \method augments PI within the tree-ensemble detector iForest \cite{conf/icdm/LiuTZ08}, our solution can easily be applied to any other ensemble based detector (\S\ref{sec:proposed}). 
	
	\item \textbf{Applications:} Besides extensive simulation experiments, we employ \method on three real-world case studies where PI respectively captures ($i$) expert knowledge, ($ii$) computationally-expensive features, and ($iii$) ``historical future'' data, which demonstrate the benefits that PI can unlock for anomaly detection in terms of 
	accuracy, speed, and detection latency (\S\ref{sec:experiments}).	
\end{itemize}

\noindent\textbf{Reproducibility:} Implementation of \method and  real world datasets used in experiments are open-sourced at \url{http://www.andrew.cmu.edu/user/shubhras/SPI}. 

\section{Motivation: {{How can LUPI benefit anomaly detection?}}} 
\label{sec:prelim}
The implications of the LUPI paradigm  for anomaly detection is particularly exciting. Here, we discuss a number of detection scenarios and demonstrate that LUPI unlocks advantages for anomaly detection problems in multiple aspects. 

In the original LUPI framework \cite{journals/nn/VapnikV09}, privileged information (hereafter PI) is defined as data that is available \textit{only} at training stage for training examples but \textit{unavailable at test time} for test examples.
Several anomaly detection scenarios admit this definition directly. 
Interestingly, PI can also be specified as \textit{strategically ``unavailable''} for anomaly detection. That is, one can willingly avoid using certain data at test time (while incorporating such data into detection models at train phase\footnote{Note that training phase in anomaly detection does not involve the use of any labels.}) in order to achieve resource efficiency, speed, and robustness. We organize detection scenarios into two with PI as (truly) Unavailable vs. Strategic, and elaborate with examples below. Table \ref{tab:benefits_of_privileged} gives a summary.

\begin{table}
	\caption{\label{tab:benefits_of_privileged} Types of data used in anomaly detection with various overhead on resources (\$\$\$, storage, battery, etc.), timeliness, and/or risk, \textit{if used as privileged information can enable resource-frugal, early, as well as preventive detection}.}
	\centering
	\begin{tabular}{l || >{\centering\arraybackslash}p{2.5cm}| >{\centering\arraybackslash}p{2cm}
			>{\centering\arraybackslash}p{1.5cm} >{\centering\arraybackslash}p{1.5cm}}
		\toprule
		\makecell{\textbf{Properties vs.} \\ \textbf{Type of Privileged Info}}                 
		& \rotatebox[origin=c]{0}{\makecell{\textbf{Unavailable }\\ \textbf{vs. Strategic}}} &  
		\rotatebox[origin=c]{0}{\makecell{\textbf{need} \\ \textbf{Resources}}} & 
		\rotatebox[origin=c]{0}{\makecell{\textbf{cause} \\\textbf{Delay}}} & 
		\rotatebox[origin=c]{0}{\makecell{\textbf{incur} \\\textbf{Risk}}} \\
		\midrule			
		1.	``historical future'' data       & U & n/a & n/a & n/a \\ 
		2. after-the-fact data				  & U & n/a & n/a & n/a \\ 
		3. 	advanced technical data & U & n/a & n/a & n/a \\ 
		\hline
		4. 	restricted-access data				& U, S & \tick &  &  \\ 
		5. 	expert knowledge 			 & U, S & \tick & \tick &  \\ 
		6. 	compute-heavy data			& S & \tick & \tick &  \\ 
		7.	unsafe-to-collect data 		& S &      & \tick &  \tick \\ 
		8.	easy-target-to-tamper data & S &  &  & \tick \\ 
		
		\bottomrule
	\end{tabular}
\end{table}

\subsubsection{Unavailable PI:} This setting includes typical scenarios, where PI is (truly) unknown for test examples. 

1. \textit{``historical future'' data}: When training an anomaly detection model with offline/historical data that is over time (e.g., temporal features), one may use values both before \textit{and after} time $t$ while creating an example for each $t$. Such data is PI; not available when the model is deployed to operate in real-time.

2. \textit{after-the-fact data}: In malware detection, the goal is to detect 
before it gets hold of and harms the system.
One may have historical data for some (training) examples from past exposures, 
including measurements of system variables (number of disk/port read/writes, CPU usage, etc.). Such after-the-exposure measurements can be incorporated as PI.

3. \textit{advanced technical data}: This includes scenarios where some (training) examples are well-understood but those to be detected are simply unknown.
For example, the expected behavior of various types of apps on a system may be common domain knowledge that can be converted to PI, but such knowledge may not (yet) be available for new-coming apps.

\subsubsection{Strategic PI:} Strategic scenarios involve PI that can in principle be acquired but is willingly avoided at test time to achieve gains in resources, time, or risk. 

4. \textit{restricted-access data}: One may want to build models that do not assume access to private data or intellectual property at test time, such as source code (for apps or executables), \textit{even if} they 
could be acquired through resources. Such information can also be truly unavailable, e.g. encrypted within the software.

5. 	\textit{expert knowledge}: Annotations about some training examples 
may be available from experts, which are truly unavailable at test time.
One could also strategically choose to avoid expert involvement at test time, which (a) may be costly to obtain and/or (b) cause significant delay, especially for real-time detection.

6. 	\textit{compute-heavy data}: One may strategically choose not to rely on features that are computationally expensive to obtain, especially in real-time detection, but rather use such data as PI (which can be extracted offline at training phase). Such features not only cause delay but also require compute resources (which e.g., may drain batteries in detecting malware apps on cellphones).

7.	\textit{unsafe-to-collect data}: This involves cases where collecting PI at test time is unsafe/dangerous. For example, the slower a drone moves to capture high-resolution (privileged) images for surveillance, not only it causes delay but more importantly, the more susceptible it becomes to be taken down.

8.	\textit{easy-target-to-tamper data}: Finally, one may want to avoid relying on features that are easy for adversaries to tamper with. Examples to those features include self-reported data (like age, location, etc.).
Such data may be available reliably for some training examples and can be used as PI.

In short, by strategically designating PI one can achieve resource, timeliness, and robustness gains for various anomaly detection tasks.
Designating features that need resources as PI $\rightarrow$ allow resource-frugal (``lazy'') detection;
features that cause delay as PI $\rightarrow$ allow early/speedy detection; and
designating features that incur vulnerability as PI $\rightarrow$ allow preventive and more robust detection.

In this subsection, we laid out a long list of scenarios that make LUPI-based learning particularly attractive for anomaly detection.
In our experiments (\S \ref{sec:experiments}) we demonstrate its premise for scenarios 1., 5. and 6. above using three real world datasets,
while leaving others as what we believe interesting future investigations.

\section{Privileged Info-Augmented Anomaly Detection}
\vspace{-0.05in}
\label{sec:proposed}
\subsubsection{The Learning Setting}
Formally, the input for the anomaly detection model at learning phase are tuples of the form

$$
\mathcal{D} = \{(\bx_1, \bxs_1), (\bx_2, \bxs_2), \ldots,  (\bx_n, \bxs_n)\}\;,
$$
where $\bx_i = (x_i^1,\ldots,x_i^d) \in X$
and $\bxs_i = (x_i^{*1},\ldots,x_i^{*p}) \in X^*$. Note that this is an unsupervised learning setting where label information, i.e., $y_i$'s are not available.
The privileged information is represented as a feature vector $\bx^* \in \mathbb{R}^p$ that is in space $X^*$,
which is \textit{additional to and different from} the feature space $X$ in which the primary information is represented as a feature vector $\bx \in \mathbb{R}^d$.

The important distinction from the traditional anomaly detection setting is that the input to the (trained) detector at testing phase are feature vectors

$$
\{\bx_{n+1}, \bx_{n+2}, \ldots, \bx_{n+m}\}\;.
$$
That is, the (future) test examples do not carry any privileged information. The anomaly detection model is to score the incoming/test examples and make decisions solely based on the primary features $\bx \in X$.

In this text, we refer to space $X^*$ as the {\em privileged space} and to $X$ as the {\em decision space}.
Here, a key assumption is that the information in the privileged space
is intelligent/nontrivial, that is, it allows to create models $f^*(\bxs)$ that detect anomalies with vectors $\bxs$ corresponding to vectors $\bx$ with higher accuracy than models $f(\bx)$. 
As a result, the main question that arises which we address in this work is:
``how can one use the knowledge of the information in space $X^*$ to improve the
performance of the desired model $f(\bx)$ in space $X$?''

In what follows, we present a first-cut attempt to the problem
that is a natural knowledge transfer between the two feature spaces (called \kt for feature transfer).
We then lay out the shortcomings of such an attempt,
and present our proposed solution \method.
We compare to \kt (and other baselines) in experiments. 

\subsection{First Attempt: Incorporating PI by Transfer of Features}
\vspace{-0.05in}
\label{ssec:ft}
A natural attempt to learning under privileged information that is unavailable for test examples
is to treat the task as a \textit{missing data problem}. Then, typical techniques for data imputation can be employed where missing (privileged) features are replaced with their predictions from the available (primary) features.

In this scheme, one simply maps vectors $\bx \in X$ into
vectors $\bx^*\in X^*$ and then builds a detector model 
in the transformed space.
The goal is to find the transformation of vectors $\bx=(x^1,\ldots,x^d)$
into vectors $\bphi(\bx) = (\phi_1(\bx),\ldots,\phi_p(\bx))$  that minimizes the expected risk given as

\begin{equation}
R(\bphi) = \sum_{j=1}^p\; \min\limits_{\phi_j} \int (x^{*j} - \phi_j(\bx))^2 p(x^{*j},\bx) dx^{*j} d\bx \;,
\end{equation}
where $p(x^{*j},\bx)$ is the joint probability of coordinate $x^{*j}$
and vector $\bx$, and functions $\phi_j(\bx)$ are defined by $p$ regressors.

Here, one could construct approximations to functions $\phi_j(\bx)$, $j=\{1,\ldots,p\}$ by solving 
$p$ regression estimation problems based on the training examples 
$$
(\bx_1,x^{*j}_1),\ldots,(\bx_n,x^{*j}_n), \;\; j = 1,\ldots,p \;,
$$
where $\bx_i$'s are input to each regression $\phi_j$ and the $j$th coordinate of the corresponding
vector $\bxs_i$, i.e. $x^{*j}_i$'s are treated as the output, by minimizing the regularized empirical loss functional

\begin{equation}
R(\phi_j) = \;\min\limits_{\phi_j} \;\; \sum_{i=1}^n  (x_i^{*j} - \phi_j(\bx_i))^2 + \lambda_j \text{penalty}(\phi_j), \;\; j = 1,\ldots,p \;.
\end{equation}

Having estimated the transfer functions $\hat{\phi}_j$'s (using linear or non-linear regression techniques), one can then learn any desired anomaly detector $f(\hat{\bphi}(\bx))$ using the training examples, which concludes the learning phase. Note that the detector does not require access to privileged features $\bxs$ and can be employed solely on primary features $\bx$ of the test examples $i=n+1,\ldots, m$.

\subsection{Proposed \method: Incorporating PI by Transfer of Decisions}
\label{ssec:spi}

Treating PI as missing data and predicting $\bxs$ from $\bx$
could be a difficult task, when privileged features are complex and high dimensional (i.e., $p$ is large).
Provided $f^*(\bxs)$ is an accurate detection model, a more direct goal would be to
\textit{mimic its decisions}---the scores that $f^*$ assigns to the training examples. Mapping \textit{data} between two spaces, as compared to \textit{decisions}, would be attempting to solve a more general problem, that is likely harder and unnecessarily wasteful. 


The general idea behind transferring decisions/knowledge (instead of data) is to identify a small number of elements in the privileged space $X^*$ that well-approximate the function $f^*(\bxs)$, and then try to transfer them to the decision space---through the approximation of those elements in space $X$. 
This is the knowledge transfer mechanism in LUPI by Vapnik and Izmailov \cite{conf/slds/VapnikI15}.
They illustrated this mechanism for the (supervised) SVM classifier. We generalize this concept to unsupervised anomaly detection.

The knowledge transfer mechanism uses three building blocks of knowledge representation in AI,
as listed in Table \ref{table:ai}.
We first review this concept for SVMs, followed by our proposed \method.
While \method is clearly different in terms of the task it is addressing as well as in its approach, as we will show, it is inspired by and builds on the same fundamental mechanism.

\begin{table}
	\caption{{\small{Three building blocks of knowledge representation in artificial intelligence, 
				in context of SVM-LUPI  for classification \cite{conf/slds/VapnikI15}
				and \method for anomaly detection [this paper].}}\label{table:ai}}
	\begin{tabular}{lp{0.25cm}rp{0.25cm}r}
		\toprule
		&& \textbf{SVM-LUPI} && \textbf{\method} (\textbf{Proposed}) \\
		\midrule
		1. Fundamental elements of knowledge && support vectors && isolation trees \\ \hline
		2. Frames (fragments) of the knowledge && kernel functions && tree anomaly scores \\\hline
		3. Structural connections of the frames && weighted sum && weighted sum (by L2R) \\
		\bottomrule
	\end{tabular}	
\end{table}

\subsubsection{Knowledge transfer for SVM:} The \textit{fundamental elements} of knowledge in the SVM classifier are the support vectors. In this scheme, one constructs two SVMs; one in $X$ space and another in $X^*$ space.
Without loss of generality, let $\bx_1,\ldots,\bx_t$ be the support vectors of SVM solution in space $X$ and $\bxs_1,\ldots,\bxs_{t^*}$
be the support vectors of SVM solution in space $X^*$, where $t$ and $t^*$ are the respective number of support vectors. 

The decision rule $f^*$ in space $X^*$ (which one aims to mimic) has the form

\begin{equation}
\label{connsvm}
f^*(\bxs) = \sum_{k=1}^{t^*} y_k \alpha_k^* K^*(\bxs_k,\bxs) + b^* \;,
\end{equation}
where $K^*(\bxs_k,\bxs)$ is the kernel function of similarity between support vector $\bxs_k$ and vector $\bxs \in X^*$,
also referred as the \textit{frames} (or {\em fragments}) of knowledge. Eq. \eqref{connsvm} depicts the \textit{structural connection} of these fragments, which is a weighted sum with learned weights $\alpha_k^*$'s.

The goal is to approximate each fragment of knowledge $K^*(\bxs_k,\bxs)$, $k = 1,\ldots, t^*$ in $X^*$
using the fragments of knowledge in $X$; i.e., the $t$ kernel functions $K(\bx_1, \bx), \ldots, K(\bx_t, \bx)$ of the SVM trained in $X$.
To this end, one maps $t$-dimensional vectors $\bz = (K(\bx_1,\bx),\ldots, K(\bx_{t},\bx)) \in Z$  into $t^*$-dimensional vectors $\bzs = (K^*(\bxs_1,\bxs),\ldots, K^*(\bxs_{t^*},\bxs)) \in Z^*$ through $t^*$ regression estimation problems. 
That is, the goal is to find regressors $\phi_1(\bz),\ldots,\phi_{t^*}(\bz)$ in $X$ such that

\begin{equation}
\label{regress}
\phi_k(\bz_i) \approx K^*(\bxs_k,\bxs_i), \;\; k=1,\ldots,t^*
\end{equation}
\vspace{-0.2in}

\noindent
for all training examples $i=1,\ldots,n$. For each $k = 1,\ldots, t^*$, one can construct the approximation to function $\phi_k$ by training a regression on the data 

$$
\{ (\bz_1, K^*(\bxs_k,\bxs_1)),\;\ldots,\;(\bz_n, K^*(\bxs_k,\bxs_n))\}, \; k = 1,\ldots, t^* \;,
$$

\noindent
where we regress vectors $\bz_i$'s onto scalar output $K^*(\bxs_k,\bxs_i)$'s to obtain $\hat{\phi}_k$.

For the prediction of a test example $\bx$, one can then replace each $K^*(\bxs_k,\bxs)$ in Eq. \eqref{connsvm} (which requires privileged features $\bxs$) with $\hat{\phi}_k(\bz)$ (which mimics it, 
using only the primary features $\bx$---to be exact, by first transforming $\bx$ into $\bz$ through the frames $K(\bx_j, \bx), j=1,\ldots,t$ in the $X$ space).




\subsubsection{Knowledge transfer for \method:} 
In contrast to mapping of features from space $X$ to space $X^*$,
knowledge transfer of decisions maps space $Z$ to $Z^*$ in which fragments of knowledge are represented. 
Next, we show how to generalize these ideas to anomaly detection with no label supervision.
Figure \ref{fig:transfer} shows an overview.

To this end, we utilize a state-of-the-art ensemble technique for anomaly detection, called  
Isolation Forest \cite{conf/icdm/LiuTZ08} (hereafters \ifr, for short), which  builds a set of extremely randomized trees. In essence, each tree approximates density in a random feature subspace and anomalousness of a point is quantified by the sum of such partial estimates across all trees. 

In this setting, one can think of the individual trees in the ensemble to constitute the {\em fundamental elements} and the partial density estimates (i.e., individual anomaly scores from trees) to constitute the {\em fragments} of knowledge, where the structural connection of the fragments is achieved by an unweighted sum.

\begin{figure}[!t]
	\vspace*{-0.15in}
	\centering
	\includegraphics[width=0.9\linewidth,height=1.4in]{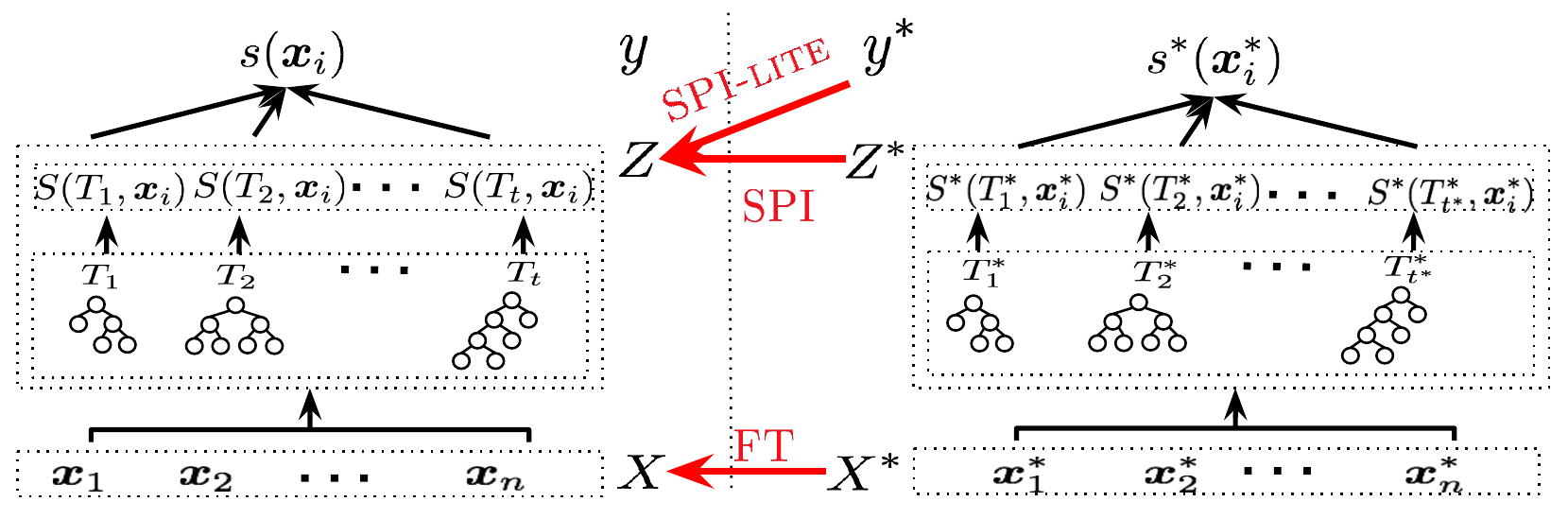}
	\vspace{-0.05in}
	\caption{Anomaly detection with PI illustrated. \kt maps \textit{data} between spaces (\S\ref{ssec:ft}) whereas \method (and ``light'' version \spiscores) mimic \textit{decisions} (\S\ref{ssec:spi}). \label{fig:transfer}}
	\vspace{-0.05in}
\end{figure}

Similar to the scheme with SVMs, we construct two {\ifr}s; one in $X$ space and another in $X^*$ space. 
Let $\mathcal{T} = T_1,\ldots,T_t$ denote the trees in the ensemble in $X$ and $\mathcal{T^*}= T^*_1,\ldots,T^*_{t^*}$
the trees in the ensemble in $X^*$, where $t$ and $t^*$ are the respective number of trees (prespecified by the user, typically a few 100s). Further, let $S^*(T^*_k,\bxs)$ denote the anomaly score estimated by tree $T^*_k$ for a given $\bxs$ (the lower the more anomalous; refer to \cite{conf/icdm/LiuTZ08} for details of the scoring).
$S(T_k,\bx)$ is defined similarly. Then, the anomaly score $s^*$ for a point $\bxs$ in space $X^*$ (which we aim to mimic) is written as
\vspace{-0.2in}
\begin{equation}
\label{sxs}
s^*(\bxs) = \sum_{k=1}^{t^*}  S^*(T^*_k,\bxs) \;,
\end{equation}
\vspace{-0.15in}

\noindent
which is analogous to Eq. \eqref{connsvm}.
To mimic/approximate each fragment of knowledge $S^*(T^*_k,\bxs)$, $k = 1,\ldots, t^*$ in $X^*$
using the fragments of knowledge in $X$; i.e., the $t$ scores for $\bx$: $S(T_1, \bx), \ldots, S(T_t, \bx)$ of the \ifr trained in $X$, we estimate $t^*$ regressors $\phi_1(\bz),\ldots,\phi_{t^*}(\bz)$ in $X$ such that

\vspace{-0.15in}
\begin{equation}
\label{regressif}
\phi_k(\bz_i) \approx S^*(T^*_k,\bxs_i), \;\; k=1,\ldots,t^*
\end{equation}
\vspace{-0.2in}

\noindent
for all training examples $i=1,\ldots,n$, where $\bz_i = (S(T_1, \bx_i),\ldots, S(T_t, \bx_i))$.
Simply put, each $\hat{\phi}_k$ is an approximate mapping of all the $t$ scores from the ensemble $\mathcal{T}$ in $X$ to an individual score (fragment of knowledge) by tree $T^*_k$ of the ensemble $\mathcal{T^*}$ in $X^*$.
In practice, we learn a mapping from the leaves rather than the trees of $\mathcal{T}$ for a more granular mapping.
Specifically, we construct vectors $\bz_i=(\bz'_{i1},\ldots,\bz'_{it})$ where
each $\bz'_{ik}$ is a size $\ell_k$ vector in which the value at index $\;\text{leaf}(T_k, \bx_i)$ is set to $S(T_k,\bx_i)$ and other entries to zero. Here, $\ell_k$ denotes the number of leaves in tree $T_k$ and $\text{leaf}(\cdot)$ returns the index of the leaf that $\bx_i$ falls into in the corresponding tree (note that $\bx_i$ belongs to exactly one leaf of any tree, since the trees partition the feature space). 

{\bf \spiscores: A ``light'' version.~} We note that instead of mimicking each individual fragment of knowledge $S^*(T^*_k,\bxs)$'s,
one could also directly mimic the ``final decision'' $s^*(\bxs)$.
To this end, we also introduce \spiscores, which estimates a {\em single} regressor $\phi(\bz_i) \approx s^*(\bxs_i)$ for $i=1,\ldots,n$ (also see Figure \ref{fig:transfer}). We compare \method and \spiscores~empirically in \S\ref{sec:experiments}.


{\bf Learning to Rank (L2R) like in $X^*$~:$\;\;$} An important challenge in learning to accurately mimic the scores $s^*$'s 
in Eq. \eqref{sxs} is to 
make sure that the regressors $\phi_k$'s are very accurate in their approximations in Eq. \eqref{regressif}.
Even then, it is hard to guarantee that the final ranking of points by $\sum_{k=1}^{t^*} \hat{\phi}_k(\bz_i)$
would reflect their ranking by $s^*(\bxs_i)$.
Our ultimate goal, after all, is to {\em mimic the ranking} of the ensemble in $X^*$ space since anomaly detection is a ranking problem at its heart.

\begin{algorithm}[!t]
	\caption{\method-{\sc Train}: Incorporating PI to Unsupervised Anomaly Detector\label{alg:spitrain}} 
	\begin{algorithmic}[1]
		\REQUIRE training examples $\{(\bx_1, \bxs_1),\ldots,  (\bx_n, \bxs_n)\}$
		\ENSURE  detection model (ensemble-of-trees) $\mathcal{T}$ in $X$ space; regressors $\hat{\phi}_k$'s, $k=1,\ldots,t^*$;
		$\bbeta$ (or $\bgamma$ for kernelized L2R)
		\STATE Learn $t^*$ isolation trees $\mathcal{T^*}= \{T^*_1,\ldots,T^*_{t^*}\}$ on $\bxs_i$'s $\; i=1,\ldots, n$
		\STATE Learn $t$ isolation trees $\mathcal{T}= \{T_1,\ldots,T_{t}\}$ on $\bx_i$'s $\; i=1,\ldots, n$
		\STATE Construct leaf score vectors $\bz_i$'s, $\; i=1,\ldots, n$, based on $\mathcal{T}$
		\FOR {{\bf each} $k=1,\ldots,t^*$}
		\STATE Learn regressor $\hat{\phi}_k$ of $\bz_i$'s onto $S^*(T^*_k,\bxs_i)$'s
		\STATE Obtain $\bbeta$ by optimizing $C$ in \eqref{l2rcost} (or $\bgamma$ for kernelized $C_{\psi}$)
		\ENDFOR
	\end{algorithmic}
\end{algorithm} 
\setlength{\textfloatsep}{-0.1in}

\begin{algorithm}
	\caption{\method-{\sc Test}: PI-Augmented Unsupervised Anomaly Detection \label{alg:spitest}} 
	\begin{algorithmic}[1]
		\REQUIRE 
		test examples $\{\bx_{n+1},\ldots,\bx_{n+m}\}$; $\mathcal{T}$, $\hat{\phi}_k$'s $k=1,\ldots,t^*$, $\bbeta$ (or $\bgamma$ if kernelized) 
		\ENSURE  estimated anomaly scores $\{s_{n+1}, \ldots, s_{n+m}\}$ for all test examples
		\FOR {{\bf each} test example $\bx_e$, $e=n+1,\ldots,n+m$}
		\STATE Construct leaf score vector $\bz_e = (\bz'_{e1},\ldots,\bz'_{et})$
		where entry in each $\bz'_{ek}$ for index $\text{leaf}(T_k,\bx_e)$ is set to  $S(T_k, \bx_e)$ and to 0 o.w., for $k=1,\ldots,t$
		\STATE Construct $\bphi_e = (\hat{\phi}_1(\bz_e), \ldots, \hat{\phi}_{t^*}(\bz_e))$
		\STATE Estimate anomaly score as $s_e = \bbeta \bphi_e^T$ (or $s_e=\sum_{l=1}^n \gamma_l K(\bphi_l,\bphi_e)$ if kernelized)
		\ENDFOR
	\end{algorithmic}
\end{algorithm}  
\setlength{\textfloatsep}{0.1in}

To this end, we set up an additional pairwise learning to rank objective as follows.
Let us denote by $\bphi_i = (\hat{\phi}_1(\bz_i), \ldots, \hat{\phi}_{t^*}(\bz_i))$ the $t^*$-dimensional vector of estimated knowledge fragments for each training example $i$.
For each pair of training examples, we create a tuple of the form $((\bphi_i,\bphi_j), p^*_{ij})$ where

\begin{equation}
\label{pstar}
p^*_{ij} = P(s^*_i < s^*_j) = \sigma(-(s^*_i - s^*_j)) \;,
\end{equation}
which is the probability that $i$ is ranked ahead of $j$ by anomalousness in $X^*$ space (recall that lower $s^*$ is more anomalous), where $\sigma(v)=1/(1+e^{-v})$ is the sigmoid function. Notice that the larger the gap between the anomaly scores of $i$ and $j$, the larger this probability gets (i.e., more surely $i$ ranks above $j$)

Given the training pair tuples above, our goal of learning-to-rank is to estimate $\bbeta \in \mathbb{R}^{t^*}$,
such that

\begin{equation}
\label{p}
p_{ij} = \sigma(\Delta_{ij}) = \sigma(\bbeta \bphi_i^T - \bbeta \bphi_j^T) 
=  \sigma(-\hat{s^*_i} + \hat{s^*_j})) \approx p^*_{ij}, \;\; \forall i,j \in \{1,\ldots,n\} \;.
\end{equation}
\vspace{-0.2in} 

We then utilize the cross entropy as our cost function over all $(i,j)$ pairs,  as

{\small{
		\begin{align}
		\label{l2rcost}
		\min\limits_{\bbeta} \; C & = \sum_{(i,j)}-p^*_{ij} \log(p_{ij}) - (1-p^*_{ij})  \log(1-p_{ij}) 
		= \sum_{(i,j)} -p^*_{ij} \Delta_{ij} + \log(1+e^{\Delta_{ij}})
		\end{align}
}}

\noindent
where $p^*_{ij}$'s are given as input to the learning as specified in Eq. \eqref{pstar} and
$p_{ij}$ is denoted in Eq. \eqref{p} and is parameterized by $\bbeta$ that is to be estimated.

The objective function in \eqref{l2rcost} is convex and can be solved via a gradient-based optimization, where
$\frac{\text{d}C}{\text{d}\mathbf{\bbeta}} = \sum_{(i,j)} (p_{ij} - p^*_{ij})(\bphi_i-\bphi_j)$ (details omitted for brevity). More importantly, in case the linear mapping $s^*_i \approx \bbeta \bphi_i^T$ is not sufficiently accurate to capture the desired pairwise rankings,
the objective can be {\em kernelized} to learn a {\em non-linear} mapping that is likely more accurate.
The idea is to write ${\bbeta_{\psi}} = \sum_{l=1}^n \gamma_l \psi(\bphi_l)$ (in the transformed space)
as a weighted linear combination of (transformed) training examples,
for feature transformation function $\psi(\cdot)$ and parameter vector
$\bgamma\in \mathbb{R}^n$ to be estimated.
Then, $\Delta_{ij}$ in objective \eqref{l2rcost} in the transformed space can be written as

\begin{equation}
\label{kernel}
\Delta_{ij} = 
\sum_{l=1}^n \gamma_l [\psi(\bphi_l)\psi(\bphi_i)^T - \psi(\bphi_l)\psi(\bphi_j)^T]
= \sum_{l=1}^n \gamma_l [K(\bphi_l,\bphi_i) - K(\bphi_l,\bphi_j)].
\end{equation}

The kernelized objective, denoted $C_{\psi}$, can also be solved through gradient-based optimization
where we can show partial derivatives (w.r.t. each $\gamma_l$) to be equal to
$\frac{\partial C_{\psi}}{\partial \gamma_l} = \sum_{(i,j)} (p_{ij} - p^*_{ij}) [K(\bphi_l,\bphi_i) - K(\bphi_l,\bphi_j)]$. 
Given the estimated $\gamma_l$'s, prediction of score is done by $\sum_{l=1}^n \gamma_l K(\bphi_l,\bphi_e)$ for any (test) example $e$.

\subsubsection{The \method Algorithm:} We outline the steps of \method for both training and testing (i.e., detection) in Algo. \ref{alg:spitrain} and Algo. \ref{alg:spitest}, respectively. Note that the test-time detection no longer relies on the availability of privileged features for the test examples,
but yet be able to leverage/incorporate them through its training.

\section{Experiments}
\label{sec:experiments}
We design experiments to evaluate our methods in two different settings:
\begin{enumerate}
	\item \textbf{Benchmark Evaluation}: We show the effectiveness of augmenting PI (see Table~\ref{table:simulation_1_pr}) on 17 publicly available benchmark datasets.\footnote{\scriptsize{\url{http://agents.fel.cvut.cz/stegodata/Loda.zip}}}
	\item \textbf{Real-world Use Cases}: We conduct experiments on LingSpam\footnote{\label{note1}\scriptsize{\url{http://csmining.org/index.php/ling-spam-datasets.html}}} and BotOrNot\footnote{\label{note2}\scriptsize{\url{https://botometer.iuni.iu.edu/bot-repository/datasets/caverlee-2011/caverlee-2011.zip}}} datasets to show that ($i$) domain-expert knowledge as PI improves spam detection, ($ii$) compute-expensive PI enables fast detection at test time, and ($iii$) ``historical future'' PI allows early detection of bots.
\end{enumerate}

\subsubsection{Baselines}
We compare both \spi{} and \spiscores{} to the following baselines:

\begin{enumerate}		
	\item \iforest ($X$-only): Isolation Forest \cite{conf/icdm/LiuTZ08} serves as a simple baseline that operates solely in decision space $X$. PI is not used neither for modeling nor detection.
	\item \textsc{\oc} (PI-incorporated): \ocsvm{} for short, is an extension of (unsupervised) One-Class SVM that incorporates PI as introduced in \cite{conf/icdm/BurnaevS16}. 
	\item \ft (PI-incorporated): This is the direct feature transfer method that incorporates PI by learning a mapping $X \rightarrow X^*$ as we introduced in \S\ref{ssec:ft}.
	\vspace*{0.05in}
	\item[\textbf{*}] \iforestpi{} ($X^*$-only): \iforest{} that operates in $X^*$ space. We report performance by \iforestpi{} only for reference, since PI is unavailable at test time.
\end{enumerate}

\subsection{Benchmark Evaluation}
\label{ss:benchmark}

The benchmark datasets do not have an explicit PI representation. Therefore, in our experiments we introduce PI as explained below.

\begin{table}[!t]
	\centering
	\caption{Mean Average Precision~(MAP) on benchmark datasets~(avg'ed over $5$ runs) for $\gamma=0.7$. Numbers in parentheses indicate rank of each algorithm on each dataset. \iforestpi{} (for reference only) reports MAP in the $X^*$ space.}
	\resizebox{\columnwidth}{!}{%
		\begin{tabular}{l@{\hskip -0.0in}r@{\hskip 0.05in}r@{\hskip 0.05in}|@{\hskip 0.05in}r|r|r|r|r@{\hskip 0.05in}||@{\hskip 0.05in}r}
			\toprule
			Datasets	& $\;p$+$d$ & $\;\;\;\;\;\;\;n$ &	\iforest & \ocsvm	&	\ft	&	\spiscores	&	\spi	&	\iforestpi\\
			\midrule
			breast-cancer	&  30  &  357   &  0.1279	(4)&	0.0935 (6)	&	0.0974 (5)	&	0.4574 (3)	&	\textBF{0.5746} (2)	&	0.6773 (1)\\
			ionosphere	&  33  &  225   &  0.0519	(4)&	\textBF{0.2914} (1)	&	0.0590 (3)	&	0.0512 (5)	&	0.0470 (6)	&	0.0905 (2)\\
			letter-recognition	&  617  &   4197  &  0.0889	(6)&	0.1473 (4)	&	0.0908 (5)	&	0.3799 (3)	&	\textBF{0.6413} (2)	&	0.9662 (1)\\
			multiple-features	&  649  &  1200   &  0.1609	(5)&	0.1271 (6)	&	0.2044 (4)	&	0.6589 (3)	&	\textBF{0.8548} (2)	&	1.0000 (1)\\
			wall-following-robot	&  24  &  2923   &  0.1946	(5)&	0.2172 (4)	&	0.1848 (6)	&	0.4331 (3)	&	\textBF{0.5987} (2)	&	0.7538 (1)\\
			cardiotocography	&  27  &  1831   &  0.2669	(5)&	0.6107 (4)	&	0.2552 (6)	&	0.6609 (3)	&	\textBF{0.6946} (2)	&	0.8081 (1)\\
			isolet	&  617  &   4497  &  0.1533	(5)&	0.1561 (4)	&	0.1303 (6)	&	0.5084 (3)	&	\textBF{0.7124} (2)	&	0.9691 (1)\\
			libras	&  90  &  216   &  0.1368	(5)&	0.4479 (4)	&	0.0585 (6)	&	0.5175 (3)	&	\textBF{0.6806} (2)	&	1.0000 (1)\\
			parkinsons	&  22  &  147   & 0.0701	(6)&	0.0964 (4)	&	0.0714 (5)	&	0.1556 (3)	&	\textBF{0.1976} (1)	&	0.1778 (2)\\
			statlog-satimage	&  36  &  3594   &  0.2108	(6)&	0.5347 (5)	&	0.5804 (4)	&	0.9167 (3)	&	\textBF{0.9480} (2)	&	0.9942 (1)\\
			gisette	&  4971 &  3500  & 0.1231	(4)&	0.0814 (6)	&	0.0977 (5)	&	0.5593 (3)	&	\textBF{0.8769} (2)	&	0.9997 (1)\\
			waveform-1	&    21 &  3304 & 0.1322	(4)&	0.1481 (3)	&	0.0841 (6)	&	0.1234 (5)	&	\textBF{0.1556} (2)	&	0.4877 (1)\\
			madelon	&   500 & 1300  &  0.7562	(5)&	0.1167 (6)	&	\textBF{0.9973} (2)	&	0.9233 (4)	&	0.9925 (3)	&	1.0000 (1)\\
			synthetic-control	& 60 &  400 & 0.3207	(6)&	0.7889 (4)	&	0.6870 (5)	&	0.8103 (3)	&	\textBF{0.8539} (2)	&	0.9889 (1)\\
			waveform-2	& 21 & 3304  &0.1271	(5)&	\textBF{0.2828} (2)	&	0.1014 (6)	&	0.1778 (3)	&	0.1772 (4)	&	0.2944 (1)\\
			statlog-vehicle	& 18 &   629 &  0.1137	(6)&	0.3146 (5)	&	0.6326 (4)	&	0.6561 (3)	&	\textBF{0.7336} (2)	&	1.0000 (1)\\
			statlog-segment	& 18 &    1320 & 0.1250	(6)&	0.2323 (4)	&	0.1868 (5)	&	0.3304 (3)	&	\textBF{0.3875} (2)	&	0.7399 (1)\\
			\midrule				
			(Average Rank)&  &     & \footnotesize{	(5.11)}&	\footnotesize{(4.23)}	&	\footnotesize{(4.88)}	&	\footnotesize{(3.29)}	&	\footnotesize{(2.35)}	&	\footnotesize{(1.11)}\\
			\bottomrule
		\end{tabular}
	}
	\label{table:simulation_1_pr}
\end{table}

\vspace{-0.15in}
\subsubsection*{Generating privileged representation.~}
For each dataset, we introduce PI by perturbing normal observations. We designate a small random fraction~($=0.1$) of $n$ normal data points as anomalies. then, we randomly select a subset of $p$ attributes and add zero-mean Gaussian noise to the designated anomalies along the selected subset of attributes with matching variances of the selected features. The $p$ selected features represent PI since anomalies stand-out in this subspace due to added noise, while the rest of the $d$ attributes represent $X$ space. Using normal observations allows us to control for features that could be used as PI. Thus we discard the actual anomalies from these datasets where PI is unknown. 

We construct $4$ versions per dataset with varying fraction~$\gamma$ of perturbed features~(PI) retained in $X^*$ space. In particular, each set has $\gamma p$ features in $X^*$, and $(1-\gamma)p + d$ features in $X$ for $\gamma \in \{0.9, 0.7, 0.5, 0.3\}$.

\subsubsection{Results}
We report the results on perturbed datasets with $\gamma=0.7$ as fraction of features retained in space $X^*$. Table~\ref{table:simulation_1_pr} reports mean Average Precision~(area under the precision-recall curve) against~$17$~datasets for different methods. The results are averaged across 5 independent runs on stratified train-test splits. 


\begin{wrapfigure}{r}{0.5\textwidth}
	\centering
	\includegraphics[width=0.45\textwidth]{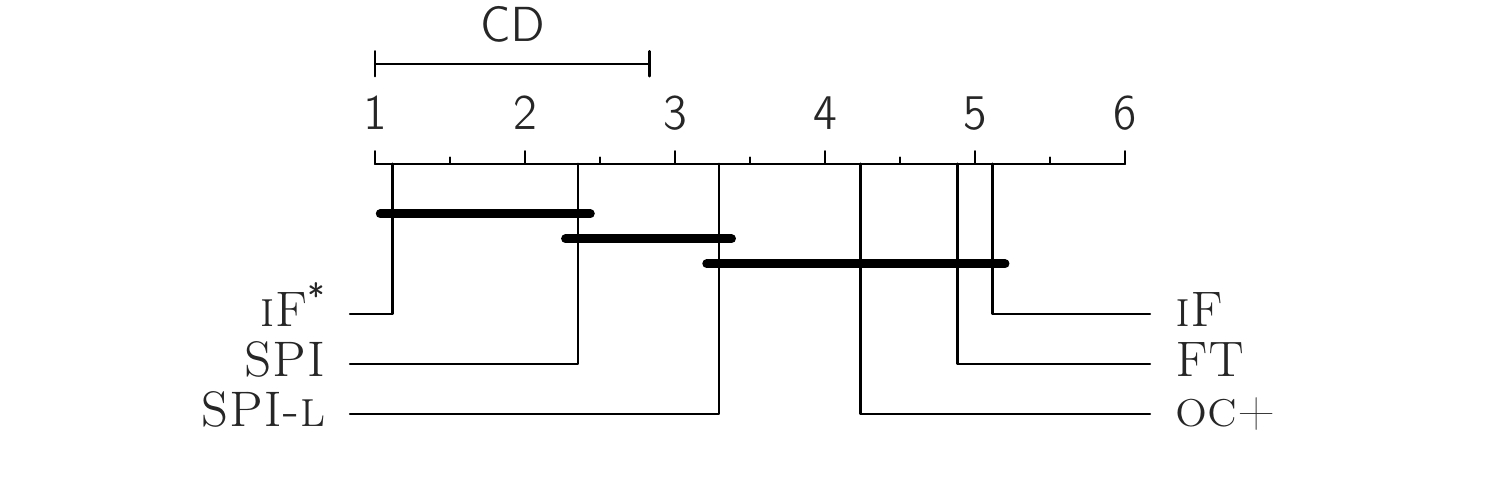}
	\caption{Average rank of algorithms (w.r.t. MAP) and comparison by the Nemenyi test. Groups of methods not significantly different (at $p\text{-val} = 0.05$) are connected with horizontal lines. CD depicts critical distance required to reject equivalence. Note that \spi{} is significantly better than the baselines.  \label{fig:cd_sim_1_diagram_0.3pr}}
	\vspace{-0.2in}
\end{wrapfigure}
Our \method outperforms competition in detection performance in most of the datasets.
To compare the methods statistically, we use the non-parametric Friedman test~\cite{demvsar2006statistical} based on the average ranks. Table~\ref{table:simulation_1_pr} reports the ranks~(in parentheses) on each dataset as well as the average ranks. With $p$-value = $2.16 \times 10^{-11}$, we reject the null hypothesis that all the methods are equivalent using Friedman test. We proceed with Nemenyi post-hoc test to  compare the algorithms pairwise and to find out the ones that differ significantly. The test identifies performance of two algorithms to be significantly different if their average ranks differ by at least the ``critical difference'' (CD). In our case, comparing 6 methods on 17 datasets at significance level $\alpha=0.05$, CD $=1.82$. 

Results of the post-hoc test are summarized through a graphical representation in Figure~\ref{fig:cd_sim_1_diagram_0.3pr}. We find that \spi{} is significant better than all the baselines. We also notice that \spi{} has no significant difference from \iforestpi{} which uses PI at test time,  demonstrating its effectiveness in augmenting PI. While all the baselines are comparable to \spiscores, its average rank is better (also see last row in Table~\ref{table:simulation_1_pr}), followed by other PI-incorporated detectors, and lastly
\iforest{} with no PI.

Average Precision (AP) is a widely-accepted metric to quantify overall performance of ranking methods like anomaly detectors.
We also report average rank of the algorithms against other popular metrics including \textsc{AUC} of ROC curve, \textsc{ndcg@10} and \textsc{precision@10} in Figure~\ref{fig:metrics_evaluation_config_1}. Notice that the results are consistent across measures, \spi{} and \spiscores{} performing among the best.

\begin{figure}[H]
	\centering
	\begin{subfigure}[b]{0.25\linewidth}
		\centering
		\includegraphics[width=\linewidth]{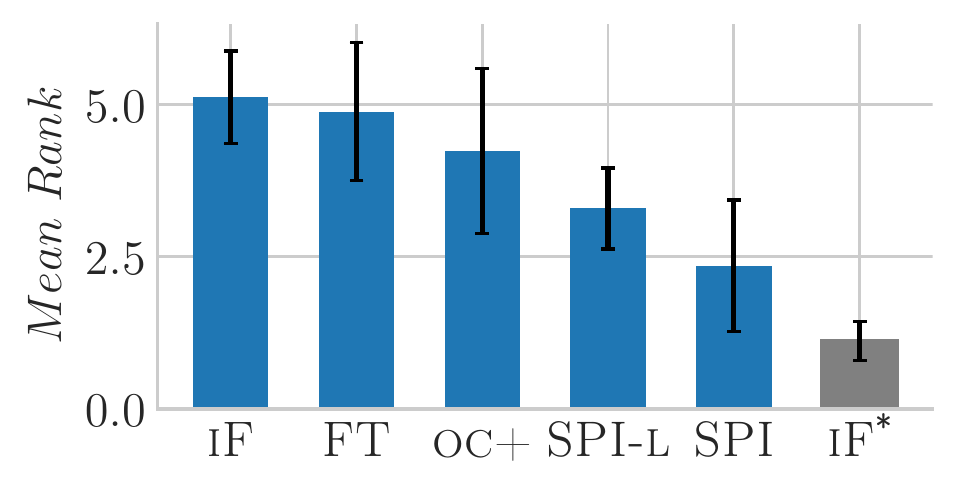}
		\caption{\textsc{MAP}}
	\end{subfigure}
	\begin{subfigure}[b]{0.24\linewidth}
		\centering
		\includegraphics[width=\linewidth]{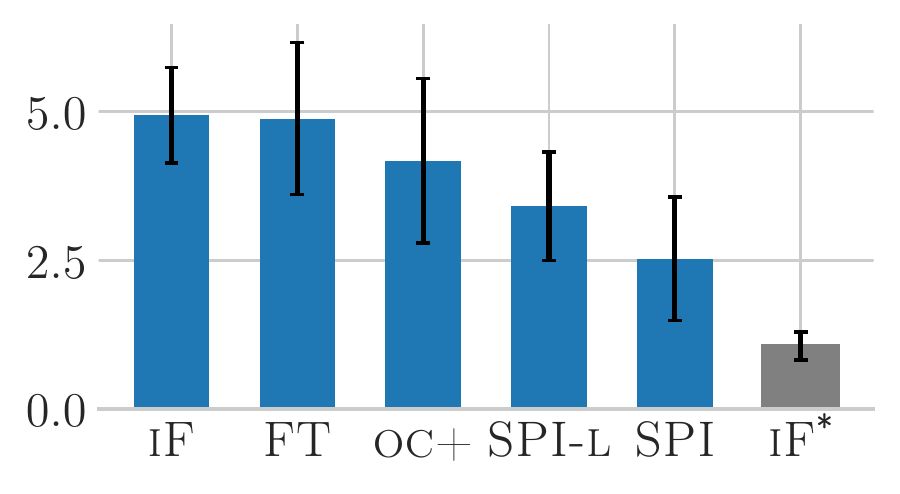}
		\caption{\textsc{AUC}}
	\end{subfigure}
	\begin{subfigure}[b]{0.24\linewidth}
		\centering
		\includegraphics[width=\linewidth]{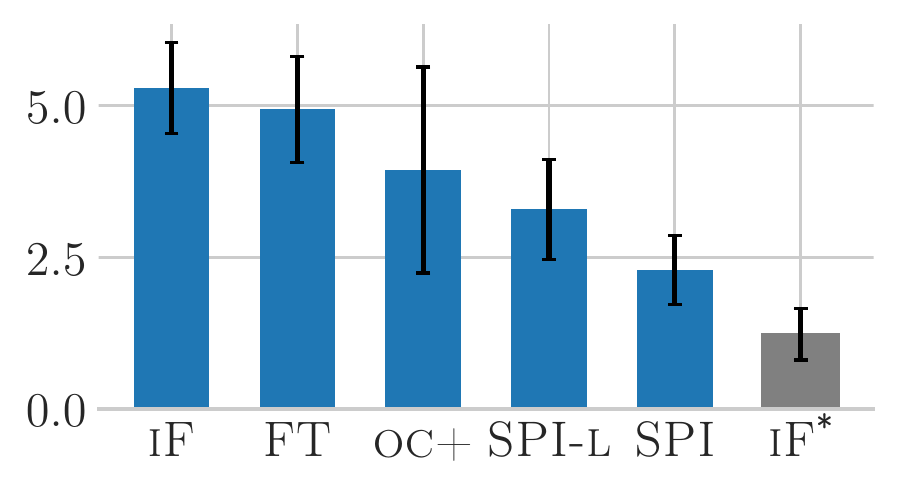}
		\caption{\textsc{ndcg@10}}
	\end{subfigure}
	\begin{subfigure}[b]{0.24\linewidth}
		\centering
		\includegraphics[width=\linewidth]{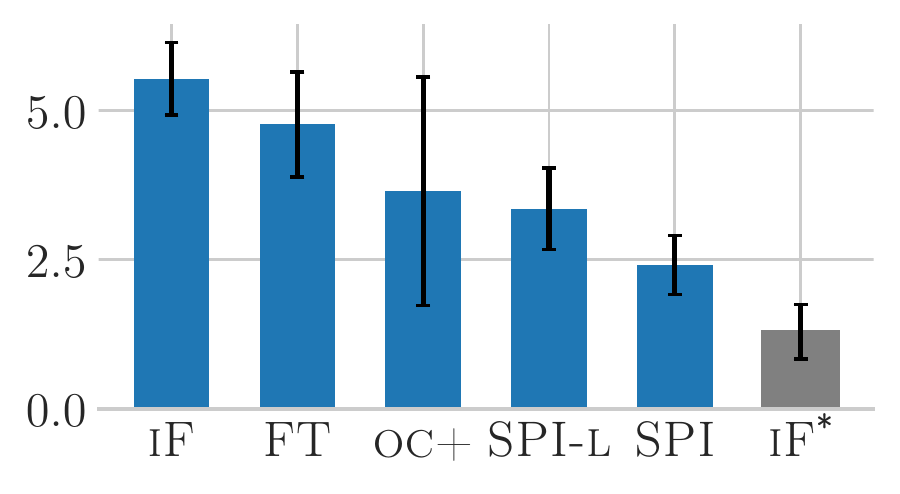}
		\caption{\textsc{precision@10}}
	\end{subfigure}
	\vspace{-.2in}
	\caption{ \method and \spil outperform competition w.r.t. different evaluation metrics.
		Average rank (bars) across benchmark datasets.   \iforestpi{} shown for reference.}
	\label{fig:metrics_evaluation_config_1}
\end{figure}

The results with $\gamma \in \{0.9, 0.5, 0.3\}$ are similar and reported in Table~\ref{table:simulation_1_pr_0.1}, Table~\ref{table:simulation_1_pr_0.5}, and Table~\ref{table:simulation_1_pr_0.7} (see Appendix \S\ref{subsec:appendix_simulation_1}).
We have performed additional simulation experiments on datasets with ground truth outliers, where original features are treated as PI and newly added columns with Gaussian noise as $X$ space (or primary) features. Details of those experiments are reported in Appendix~\S\ref{subsec:appendix_simulation_2}.

\vspace{-0.1in}

\subsection{Real-world Use Cases}

\textbf{Data description.~} LingSpam dataset\footnoteref{note1}
consists of 2412 non-spam and 481 spam email messages from a linguistics mailing-list. We evaluate two use cases (1) domain-expert knowledge as PI and (2) compute-expensive PI on LingSpam.

BotOrNot dataset\footnoteref{note2}
is collected from Twitter during December 30, 2009 to August 2, 2010. It contains 22,223 content polluters~(bots) and 19,276 legitimate users, along with their number of followings over time and tweets. For our experiments, we select accounts with age less than 10 days~(for early detection task) at the beginning of dataset collection. The subset contains 901 legitimate~(human) accounts and 4535 bots. We create 10 sets containing all the legitimate and a random 10\% sample of the bots. We evaluate use case (3) ``historical future'' as PI report results averaged over these sets.

\subsubsection*{Case 1: Domain-expert Knowledge as PI for Email Spam Detection.\\} 

\hspace*{4.5mm} $\boldsymbol{X}^*$ {\bf space:} The Linguistic Inquiry and Word Count
(LIWC) software\footnote{\url{https://liwc.wpengine.com/}} is a widely used text analysis tool in social sciences. It uses a manually-curated keyword dictionary to categorize text into 90 psycholinguistic classes. Construction of LIWC dictionary relies exclusively on human experts which is a slow and evolving process. For the LingSpam dataset, we use 
the percentage of word counts in each class~(assigned by LIWC software) as the privileged features.

$\boldsymbol{X}$ \textbf{space:} The bag-of-word model is widely used as feature representation in text analysis. As such, we use the term frequencies for our email corpus as the primary features.

\begin{wrapfigure}{r}{0.51\textwidth}
	\centering
	\includegraphics[width=0.475\textwidth]{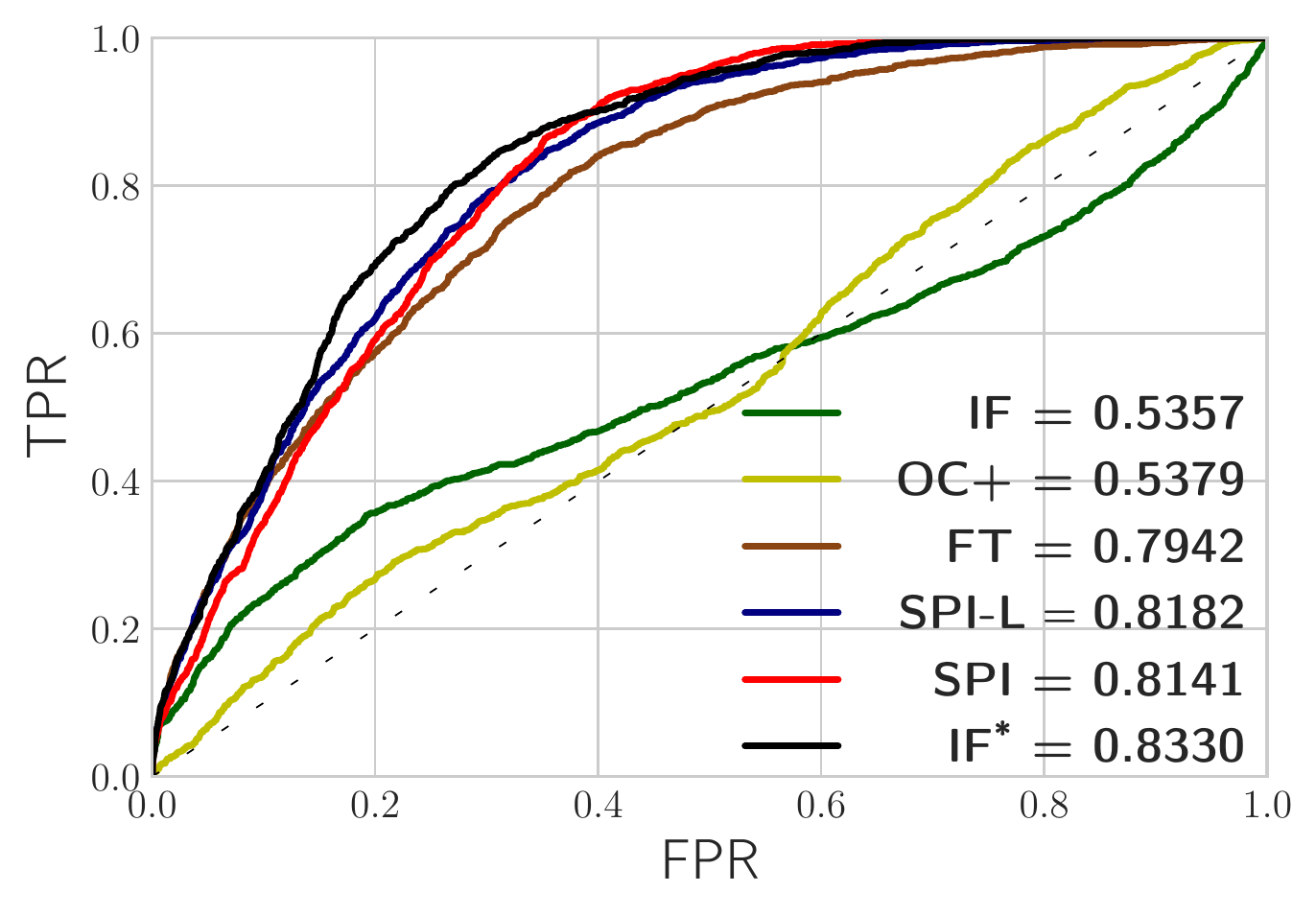}
	\vspace{-0.15in}
	\caption{\small \sloppy{Detection performance  on Case 1: using expert knowledge as PI. Legend depicts the \textsc{AUC} values. PI-incorporated detectors (except \oc) outperform non-PI \ifr and achieve similar performance to \iforestpi.} \label{fig:expert-knowledge}}
	\vspace{-0.275in}
\end{wrapfigure}
Figure~\ref{fig:expert-knowledge} shows the detection performance of algorithms in ROC curves (averaged over 15 independent runs on stratified train-test splits).
We find that \ifr, which does not leverage PI but operates solely in $X$ space, is significantly worse than most PI-incorporated methods.
\oc is nearly as poor as \ifr despite using PI---this is potentially due to OC-SVM being a poor anomaly detector in the first place, as shown in \cite{Emmott2013} and as we argued in \S\ref{sec:intro}.
All knowledge transfer methods, \method, \spil, and \kt, perform similarly on this case study, and are as good as \iforestpi, directly using $X^*$.

In Table~\ref{tab:all_metrics_real_use_case}, we report detection performance of algorithms with respect to widely used ranking metrics including \textsc{MAP}, \textsc{ndcg@k} and \textsc{precision@k} (averaged over 15 independent runs on stratified train-test splits) for use cases on LingSpam dataset.
\begin{table}[!t]
	\centering
	\caption{\spi{} and \spiscores{} rank better than competing methods w.r.t. different evaluation metrics. Results are averaged across 15 runs. \iforestpi{} shown for reference.\label{tab:all_metrics_real_use_case}}
	\resizebox{\columnwidth}{!}{%
		\begin{tabular}{c|@{\hskip 0.05in}l|@{\hskip 0.05in}rrrrr@{\hskip 0.05in}|@{\hskip 0.05in}r}
			\toprule
			Use case              & Metric & \iforest & \ocsvm	&	\ft	&	\spi\textsc{-l}	&	\spi	&	\iforestpi \\
			\midrule
			\multirow{ 3}{*}{\parbox{2.5cm}{\centering Domain-expert Knowledge as PI}} & \textsc{MAP}    & 0.2767 & 0.2021 & 0.4449 & 0.4604 & 0.4317   & 0.4708  \\
			& \textsc{AUC}    & 0.5357& 0.5379  & 0.7942  & 0.8182 & 0.8141  & 0.8330 \\
			& \textsc{ndcg@k}   & 0.3575 & 0.2468 & 0.4783 & 0.4873 & 0.4613   & 0.4975  \\
			& \textsc{p@k}   & 0.2931 & 0.2236 & 0.4265 & 0.4279 & 0.4071   & 0.4473  \\
			\midrule
			\multirow{ 3}{*}{\parbox{2.5cm}{\centering Compute-Expensive Features as PI}} & \textsc{MAP}    & 0.2767 & 0.1975 & 0.2892 & 0.3140 & 0.3131   & 0.3216  \\
			
			& AUC     & 0.5357 & 0.5369 & 0.5781 & 0.6232 & 0.6184   & 0.6572 \\
			& \textsc{ndcg@k}   & 0.3575 & 0.2275 & 0.3680 & 0.3899 & 0.3924   & 0.3930  \\
			& \textsc{p@k}   & 0.2931 & 0.2200 & 0.2974 & 0.3182 & 0.3225   & 0.3283  \\ 
			\bottomrule
		\end{tabular}
	}
\end{table}
Notice that the results are consistent~(with \textsc{AUC}) across measures, \spi{} and \spiscores{} performing among the best.

\subsubsection*{Case 2: Compute-Expensive Features as PI for Email Spam Detection.\\} 

\hspace*{4.5mm} $\boldsymbol{X}^*$ {\bf space:} 
Beyond bag-of-words, one can 
use syntactic features to capture \textit{stylistic} differences between spam and non-spam emails. 
To this end, we extract features from the parse trees of emails using the StanfordParser\footnote{\scriptsize{\url{https://nlp.stanford.edu/software/lex-parser.shtml}}}. The parser provides the taxonomy (tree) of Part-of-Speech~(PoS) tags for each sentence, based on which we construct ($i$) PoS bi-gram frequencies,
and ($ii$) quantitative features (width, height, and horizontal/vertical imbalance) of the parse tree. 

On average, StanfordParser requires 66 seconds\footnote{
	\scriptsize{using a single thread on 2.2 GHz Intel Core i7 CPU with 8 cores and 16GB RAM}}  to parse and extract features from a single raw email in LingSpam. Since the features are computationally demanding, we incorporate those as PI to facilitate faster detection at test time. 

$\boldsymbol{X}$ \textbf{space:} We use the term frequencies as the primary features as in Case 1.

Figure~\ref{fig:complex-features} (a) shows the detection performance of methods in terms of \textsc{AUC} under \textsc{ROC}. 
We find that \iforestpi using (privileged) syntactic features achieves lower AUC of $\sim$0.65 as compared to $\sim$0.83 using (privileged) LIWC features in Case 1. Accordingly, all methods perform relatively lower, suggesting that the syntactic features are less informative of spam than psycholinguistic ones.
Nonetheless, we observe that the performance ordering remains consistent, where \ifr ranks at the bottom and \method and \spil get closest to \iforestpi.


\begin{figure}[!t]
	\centering
	\vspace*{-.1in}	
	\begin{tabular}{cc}
		\includegraphics[width=0.45\linewidth]{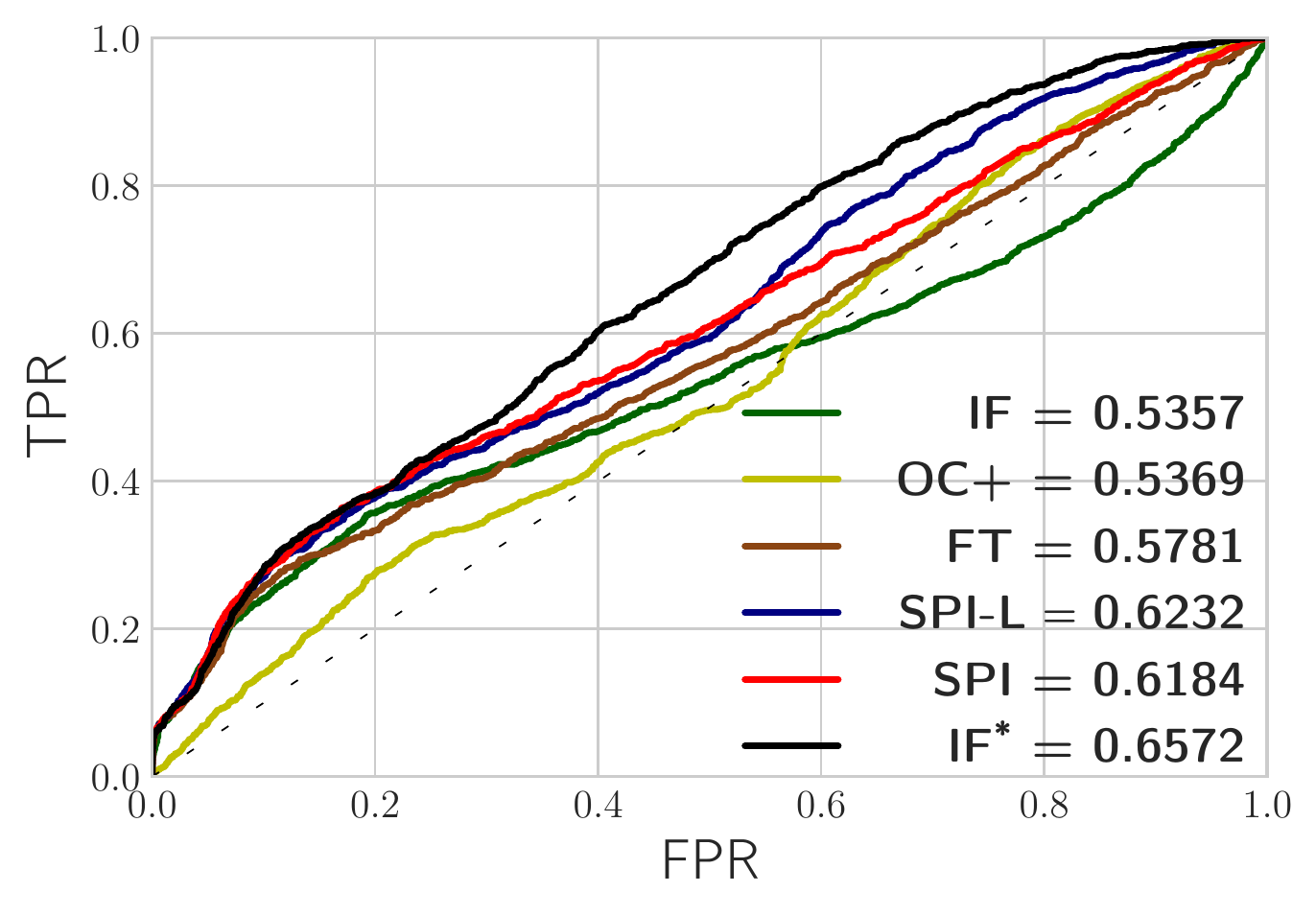} &
		\includegraphics[width=0.55\linewidth, height=1.55in]{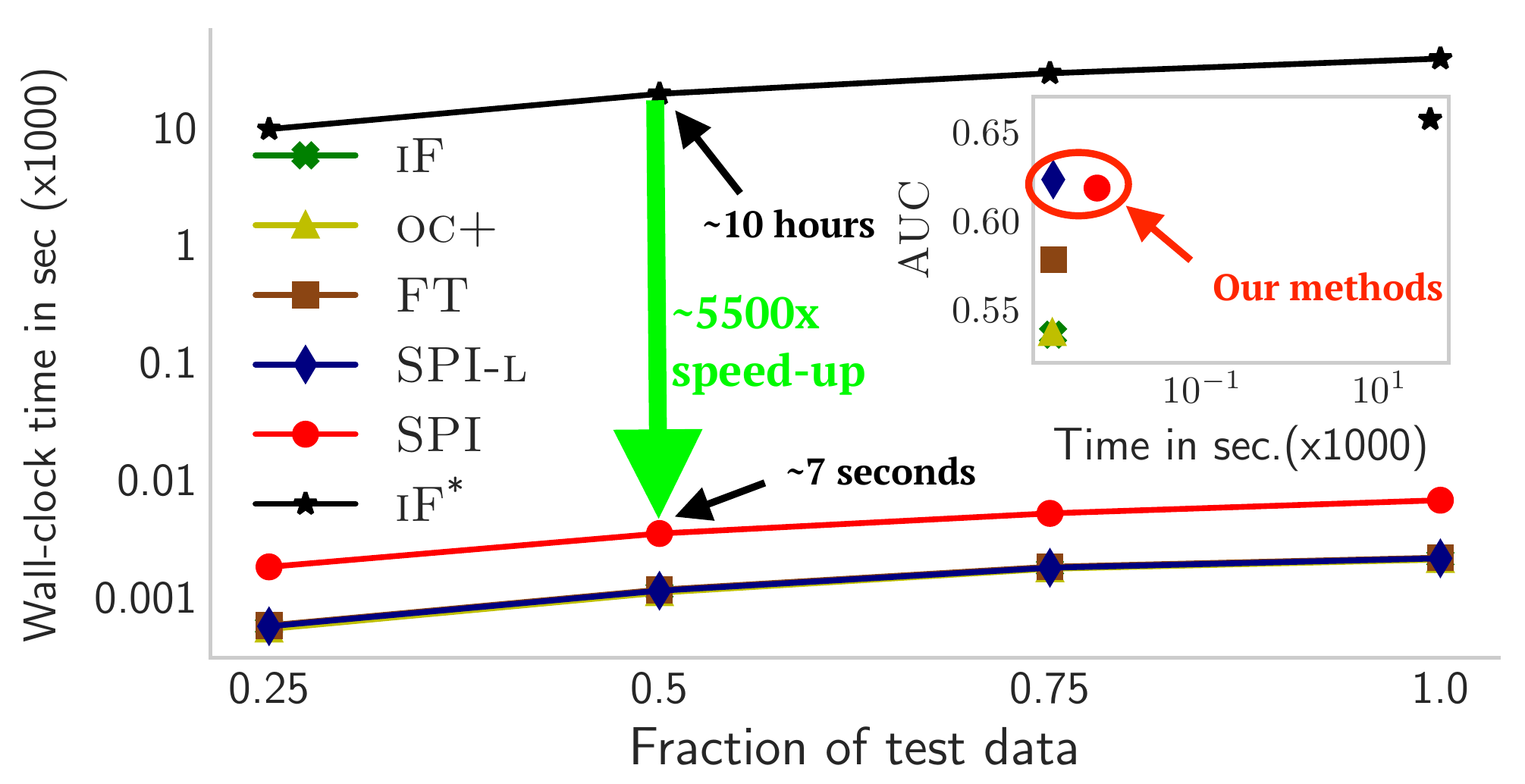}\\
	\end{tabular}
	\vspace*{-0.15in}
	\caption{\small Comparison of detectors on Case 2: using computationally-expensive features as PI. (a) detection performance, legend depicts AUC values; and 
		(b) wall-clock time required (in seconds, note the logarithmic scale) vs. test data size [inset plot on top right: AUC vs. time (methods depicted with symbols)].
		\label{fig:complex-features}}
\end{figure}

Figure~\ref{fig:complex-features} (b) shows the comparison of wall-clock time 
required by each detector to compute the anomaly scores at test time for  varying fraction of test data.
On average, \method achieves $5500\times$ speed-up over \iforestpi{} that employs the parser at test time.
This is a considerable improvement of response time for comparable accuracy. Also notice the inset plot showing the AUC vs. total test time, where our proposed \method and \spil are closest to the ideal point at the top left.


Table~\ref{tab:all_metrics_real_use_case} quantifies performance of methods against other metrics which are consistent with AUC.

\subsubsection*{Case 3: ``Historical Future'' as PI for Twitter Bot Detection.\\} 

We use temporal data from the activity and network evolution of an account to capture behavioral differences between a human and a bot. We construct temporal features including volume, rate-of-change, and lag-autocorrelations of the number of followings. We also extract temporal features from text such as count of tweets, links, hash-tags and mentions.

$\boldsymbol{X}^*$ \textbf{space}: All the temporal features within $f_t$ days in the future (relative to detection at time $t$) constitute privileged features. Such future values would not be available at any test time point but can be found in historical data.

$\boldsymbol{X}$ \textbf{space}: Temporal features within $h_t$ days in the past as well as 
static user features (from screen name and profile description) constitute primary features.


Figure~\ref{fig:historical-future} (a) reports the detection performance of algorithms in terms of ROC curves~(averaged over 10 sets) at time $t=2$ days after the data collection started; for $h_t=2$, $f_t=7$.
The findings are similar to other cases: \spi{} and \spiscores{} outperform the competing methods in terms of \textsc{AUC} and \oc{} performs similar to non-PI \iforest{}; demonstrating that knowledge transfer based methods are more suitable for real-world use cases.
\begin{figure}
	\centering
	\begin{tabular}{cc}
		\includegraphics[width=0.44\linewidth]{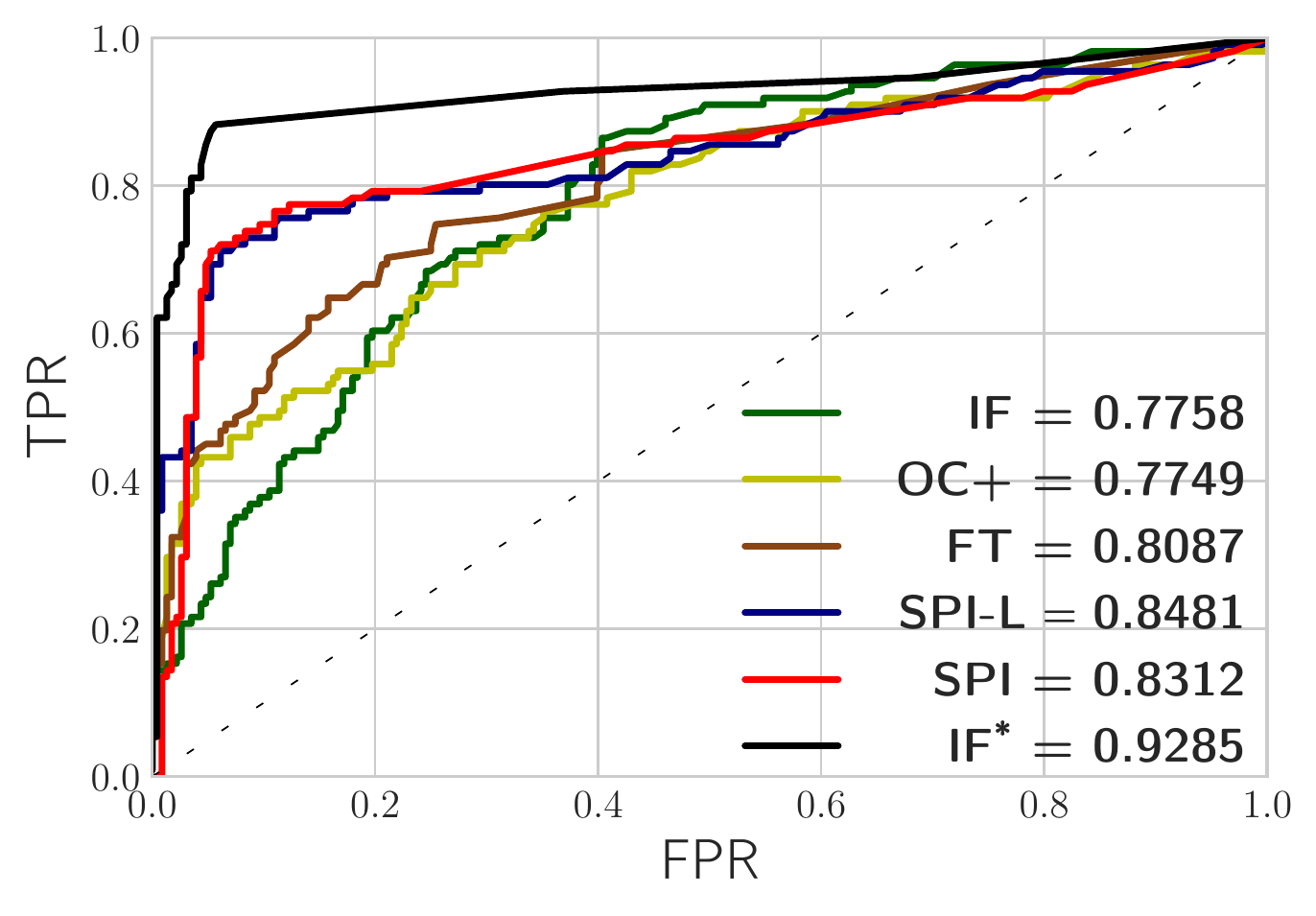} &
		\includegraphics[width=0.47\linewidth]{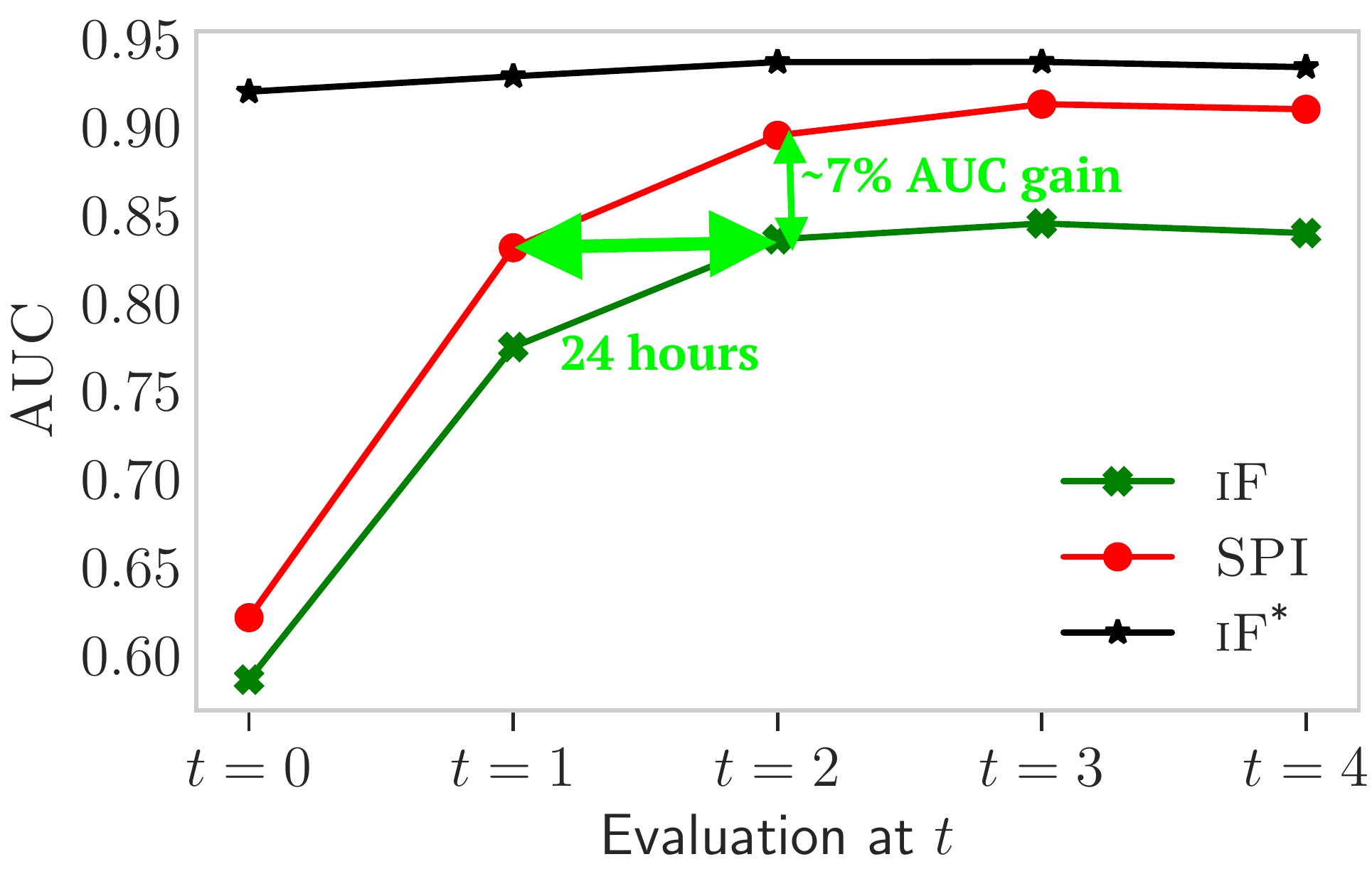}\\
	\end{tabular}
	\vspace*{-0.15in}
	\caption{\small Comparison of detectors on Case 3: using ``historical future'' data as PI. 
		(a) \method outperforms competition in performance and is closest to \iforestpi's; (b)
		\method achieves same detection performance as \ifr 24 hours earlier, and gets close to \iforestpi in 3 days of history.
		\label{fig:historical-future}}
\end{figure}

Figure~\ref{fig:steps_ahead} shows the effect of number look ahead days as PI on the performance of PI incorporated methods. We find that the gain in performance through PI saturates in about a week indicating that temporally local network changes are more valuable for bot detection task. Once again, \spi{} and \spiscores{} outperform competing methods at bot detection task over varying number of future days as PI.

\begin{figure}[!h]
	\centering
	\includegraphics[width=0.8\linewidth]{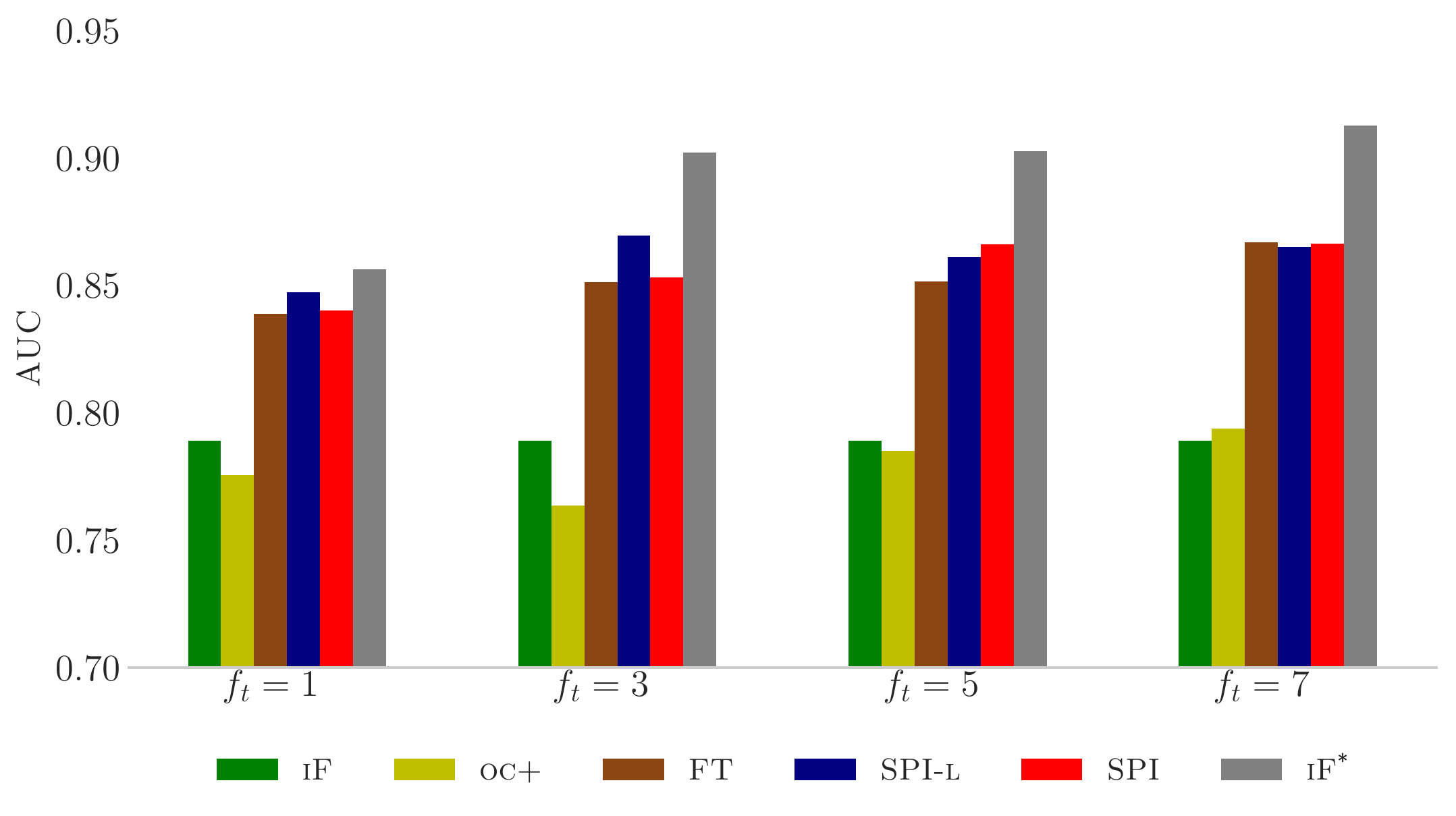}
	\caption{Effect of future information on PI incorporated method. $h_t =3$ and $f_t \in \{1, 3, 5, 7\}$ days\label{fig:steps_ahead}}
\end{figure}

Figure~\ref{fig:historical-future} (b) compares the detection performance of \spi{} and \iforest{} over time; for detection at $t = \{0, 1, 2, 3, 4\}$. As time passes, historical data grows as $h_t = \{0, 1, 2, 3, 4\}$ where ``historical future'' data is fixed at $f_t = 7$ for PI-incorporated methods. Notice that at time $t=1$, \spi{} achieves similar detection performance  to \ifr's performance at $t=2$ that uses more historical data of 2 days. As such, \spi{} enables 24 hours early detection as compared to non-PI \iforest{} for the same accuracy. Notice that with the increase in historical data, the performances of both methods improve, as expected. At the same time, that of \method improves faster, ultimately reaching a higher saturation level, specifically $\sim$7\% higher relative to \ifr. Moreover, \method gets close to  \iforestpi's level in just around 3 days. 

\section{Related Work}
\label{sec:relatedwork}

We review the history of LUPI, follow up and related work on learning with side/hidden information, as well as LUPI-based anomaly detection. 

\textbf{Learning Under Privileged Information:}
The LUPI paradigm
is introduced by Vapnik and Vashist \cite{journals/nn/VapnikV09} as the SVM+ method,
where, Teacher provides Student not only with (training) examples but also explanations, comparisons, metaphors, etc. which accelerate the learning process. Roughly speaking, PI adjusts Student's concept of similarity between training examples and 
reduces the amount of data required for learning.
Lapin et al. \cite{journals/nn/LapinHS14} showed that learning with PI is a particular instance of importance weighting in SVMs.
Another such mechanism was introduced more recently by Vapnik and Izmailov \cite{conf/slds/VapnikI15},  where knowledge is transferred from the space of PI to the space where the decision function is built.
The general idea is to specify a small number of 
fundamental concepts of knowledge 
in the privileged space
and then try to transfer them; i.e., construct additional features in decision space via e.g., regression techniques in decision space.
Importantly, the knowledge transfer mechanism is not restricted to SVMs, but generalizes, e.g. to 
neural networks \cite{journals/amai/VapnikI17}.

LUPI has been applied to a number of different settings including clustering \cite{journals/isci/FeyereislA12,conf/icpr/MarcaciniDHR14}, metric learning \cite{journals/tnn/FouadTRS13}, learning to rank \cite{conf/iccv/SharmanskaQL13}, malware and bot detection \cite{conf/icdm/BurnaevS16,celik2016extending}, risk modeling \cite{journals/eswa/RibeiroSCVN12},
as well as recognizing objects \cite{DBLP:journals/corr/SharmanskaQL14}, actions and events \cite{journals/ijcv/NiuLX16}.


\textit{Learning with Side/Hidden Information:} Several other work, particularly in computer vision 
\cite{conf/accv/ChenLL12,conf/cvpr/WangJ15a},
propose methods to learn with data that is unavailable at test time referred 
as side and hidden information  (e.g., text descriptions or tags for general images, facial expression annotations for face images, etc.).
In addition,
Jonschkowski et al. \cite{jonschkowski2015patterns} describe various patterns of learning with side information.
All of these work focus on supervised learning problems.


\textbf{LUPI-based Anomaly Detection:}
With the exception of One-Class SVM (\oc) \cite{conf/icdm/BurnaevS16}, which is a direct extension of Vapnik's (supervised) SVM+, the LUPI framework has been utilized only for supervised learning problems.
While anomaly detection has been studied extensively \cite{Aggarwal13},
we are unaware of any work other than \cite{conf/icdm/BurnaevS16} leveraging privileged information for unsupervised anomaly detection.
As we argued in \S \ref{sec:prelim} and empirically demonstrated through benchmark experiments \cite{Emmott2013}, 
\oc is not an effective solution to anomaly detection.
Motivated by this along with the premises of the LUPI paradigm,
we are the first to design a new technique that ties LUPI with unsupervised tree-based ensemble methods,
which are considered state-of-the-art for anomaly detection.

\section{Conclusion}
\label{sec:conclude}
We introduced \spi{}, a new ensemble approach that leverages privileged information~(data available only for training examples) for unsupervised anomaly detection. 
Our work builds on the LUPI paradigm, and to the best of our knowledge, is the first attempt to incorporating PI to improve the state-of-the-art {ensemble detectors}. 
We validated the effectiveness of our method on both benchmark datasets as well as three real-world case studies.
We showed that \spi{} and \spiscores{} consistently outperform the baselines.
Our case studies leveraged a variety of privileged information---``historical future'',  complex features, expert knowledge---and  verified that \method can unlock multiple benefits for anomaly detection in terms of detection latency, speed, as well as accuracy.

\hide{
\section*{Acknowledgments}\\
Any opinions, findings, and conclusions or recommendations expressed in this
material are those of the author(s) and do not necessarily reflect the views of
any of the funding parties.
}

{
		\vspace{-0.1in}
		\bibliographystyle{abbrv}
		\bibliography{refs}
}

\appendix
\section{Appendix}
\label{sec:appendix}
\subsection{Benchmark Evaluation}
\label{subsec:appendix_simulation_1}
We report the results on perturbed datasets with $\gamma \in \{0.9,0.5,0.3\}$ as fraction of features retained in space $X^*$. Table~\ref{table:simulation_1_pr_0.1}, Table~\ref{table:simulation_1_pr_0.5}, and Table~\ref{table:simulation_1_pr_0.7} report \textsc{MAP} against 17 datasets for different methods for $\gamma \in \{0.9,0.5,0.3\}$. The results are averaged across 5 independent runs on stratified train-test splits.

\begin{table}
	\centering
	\caption{Mean Average Precision (MAP) on benchmark datasets (avg'ed over 5 runs) for $\gamma=0.9$. \iforestpi{} (for reference only) reports MAP in the $X^*$ space.}
	\resizebox{\columnwidth}{!}{%
		\begin{tabular}{l@{\hskip -0.1in}rr@{\hskip 0.1in}|@{\hskip 0.1in}rcrcr@{\hskip 0.1in} |@{\hskip 0.1in}r}
			\toprule
			Dataset	& $p$$+$$d$ & $n$ &	\iforest & \ocsvm	&	\ft	&	\spiscores	&	\spi	&	\iforestpi\\
			\midrule
			breast-cancer	&  30  &  357   &  0.0921	&	0.0945	&	0.0758	&	0.0917	&	\textBF{0.1074}	&	0.6584\\
			ionosphere	&  33  &  225   &  0.0509	&	0.1010	&	0.0566	&	0.0400	&	\textBF{0.1029}	&	0.1997\\
			letter-recognition	&  617  &   4197  &  0.0756	&	0.1052	&	0.0854	&	0.0848	&	\textBF{0.3014}	&	0.9331\\
			multiple-features	&  649  &  1200   &  0.1351	&	0.1047	&	0.3521	&	0.4188	&	\textBF{0.6687}	&	1.0000\\
			wall-following-robot	&  24  &  2923   &  \textBF{0.1496}	&	0.1345	&	0.1288	&	0.1213	&	0.1223	&	0.7288\\
			cardiotocography	&  27  &  1831   &  0.1868	&	0.3577	&	0.1768	&	0.3766	&	\textBF{0.4269}	&	0.8131\\
			isolet	&  617  &   4497  &  0.1284	&	0.1251	&	0.1282	&	0.1789	&	\textBF{0.2766}	&	0.9688\\
			libras	&  90  &  216   &  0.0712	&	0.1126	&	0.0883	&	\textBF{0.2756}	&	0.1476	&	1.0000\\
			parkinsons	&  22  &  147   & 0.0443	&	0.1378	&	0.0486	&	0.0619	&	\textBF{0.1485}	&	0.1889\\
			statlog-satimage	&  36  &  3594   &  0.1387	&	0.0906	&	\textBF{0.1423}	&	0.1332	&	0.1234	&	0.9999\\
			gisette	&  4971 &  3500  & 0.0892	&	0.0843	&	0.1039	&	0.1953	&	\textBF{0.3771}	&	0.9968\\
			waveform-1	& 21 &  3304 & 0.1052	&	\textBF{0.1086}	&	0.0933	&	0.1035	&	0.0974	&	0.5201\\
			madelon	&   500 & 1300  &  0.3009	&	0.1090	&	0.5529	&	0.6215	&	\textBF{0.8232}	&	1.0000\\
			synthetic-control-chart	& 60 &  400 & 0.2495	&	0.4334	&	0.4167	&	0.3889	&	\textBF{0.5025}	&	1.0000\\
			waveform-2	& 21 & 3304  &0.0906	&	\textBF{0.1199}	&	0.0937	&	0.1105	&	0.1027	&	0.4253\\
			statlog-vehicle	& 18 &   629 &  0.0918	&	0.1309	&	0.0950	&	0.1110	&	\textBF{0.1474}	&	1.0000\\
			statlog-segment	& 18 &    1320 & 0.0997	&	0.1348	&	0.1928	&	\textBF{0.1968}	&	0.1831	&	0.7943\\
			\bottomrule
		\end{tabular}
	}
	\label{table:simulation_1_pr_0.1}
\end{table}

\begin{table}
	\centering
	\caption{Mean Average Precision (MAP) on benchmark datasets (avg'ed over 5 runs) for $\gamma=0.5$. \iforestpi{} (for reference only) reports MAP in the $X^*$ space.}
	\resizebox{\columnwidth}{!}{%
		\begin{tabular}{lrr@{\hskip 0.1in}|@{\hskip 0.1in}rcrcr@{\hskip 0.1in} |@{\hskip 0.1in}r}
			\toprule
			Dataset	& $p$$+$$d$ & $n$ &	\iforest & \ocsvm	&	\ft	&	\spiscores	&	\spi	&	\iforestpi\\
			\midrule
			breast-cancer	&  30  &  357   &  0.1718	&	0.2991	&	0.5926	&	0.5725	&	\textBF{0.6787}	&	0.8224\\
			ionosphere	&  33  &  225   &  0.0518	&	\textBF{0.1445}	&	0.0557	&	0.0428	&	0.0500	&	0.5833\\
			letter-recognition	&  617  &   4197  &  0.1127	&	0.2436	&	0.0793	&	0.4652	&	\textBF{0.7950}	&	0.9079\\
			multiple-features	&  649  &  1200   &  0.2137	&	0.1184	&	0.4486	&	0.8456	&	\textBF{0.9580}	&	1.0000\\
			wall-following-robot	&  24  &  2923   &  0.2995	&	\textBF{0.5545}	&	0.2164	&	0.4134	&	0.5035	&	0.5611\\
			cardiotocography	&  27  &  1831   &  0.4044	&	0.5945	&	0.2133	&	\textBF{0.6840}	&	0.6396	&	0.7824\\
			isolet	&  617  &   4497  &  0.1833	&	0.2360	&	0.1229	&	0.5454	&	\textBF{0.8015}	&	0.9051\\
			libras	&  90  &  216   &  0.1448	&	0.5419	&	0.8333	&	\textBF{1.0000}	&	\textBF{1.0000}	&	1.0000\\
			parkinsons	&  22  &  147   &  0.0772	&	0.0957	&	0.1369	&	0.1476	&	\textBF{0.1889}	&	0.3333\\
			statlog-satimage	&  36  &  3594   &  0.3410	&	0.6395	&	0.8667	&	0.9739	&	\textBF{0.9878}	&	0.9885\\
			gisette	&  4971 &  3500  &  0.1522	&	0.0910	&	0.1058	&	0.6013	&	\textBF{0.9494}	&	0.9921\\
			waveform-1	& 21 &  3304 &  0.1702	&	\textBF{0.4247}	&	0.1668	&	0.2952	&	0.3123	&	0.6266\\
			madelon	&   500 & 1300  &  0.9943	&	0.1115	&	\textBF{1.0000}	&	0.9811	&	0.9786	&	1.0000\\
			synthetic-control-chart	& 60 &  400 &  0.4901	&	\textBF{1.0000}	&	0.5108	&	0.8711	&	0.8875	&	0.9587\\
			waveform-2	& 21 & 3304  &  0.1881	&	\textBF{0.4368}	&	0.3215	&	0.3573	&	0.3595	&	0.4162\\
			statlog-vehicle	& 18 &   629 &  0.2397	&	0.4683	&	\textBF{0.8672}	&	0.7304	&	0.7886	&	0.8593\\
			statlog-segment	& 18 &    1320 &  0.2013	&	0.3572	&	0.1043	&	0.2983	&	\textBF{0.4190}	&	0.3724\\
			\bottomrule
		\end{tabular}
	}
	\label{table:simulation_1_pr_0.5}
\end{table}

\begin{table}
	\centering
	\caption{Mean Average Precision (MAP) on benchmark datasets (avg'ed over 5 runs) for $\gamma=0.3$. \iforestpi{} (for reference only) reports MAP in the $X^*$ space.}
	\resizebox{\columnwidth}{!}{%
		\begin{tabular}{lrr@{\hskip 0.1in}|@{\hskip 0.1in}rcrcr@{\hskip 0.1in} |@{\hskip 0.1in}r}
			\toprule
			Dataset	&$p$$+$$d$ & $n$ &	\iforest & \ocsvm	&	\ft	&	\spiscores	&	\spi	&	\iforestpi\\
			\midrule
			breast-cancer	&  30  &  357   &  0.2358	&	\textBF{0.4469}	&	0.2432	&	0.3988	&	0.4424	&	0.6506\\
			ionosphere	&  33  &  225   &  0.0534	&	\textBF{0.2382}	&	0.0482	&	0.0430	&	0.0519	&	0.0520\\
			letter-recognition	&  617  &   4197  &  0.1449	&	0.4707	&	0.0835	&	0.5839	&	\textBF{0.8361}	&	0.9382\\
			multiple-features	&  649  &  1200   &  0.3053	&	0.1238	&	0.3795	&	0.7876	&	\textBF{0.9263}	&	0.9958\\
			wall-following-robot	&  24  &  2923   &  \textBF{0.4448}	&	0.3954	&	0.3217	&	0.3528	&	0.3956	&	0.4500\\
			cardiotocography	&  27  &  1831   &  0.5827	&	\textBF{0.8330}	&	0.2060	&	0.7850	&	0.7695	&	0.8812\\
			isolet	&  617  &   4497  &  0.2753	&	0.4870	&	0.1021	&	0.6902	&	\textBF{0.8756}	&	0.8871\\
			libras	&  90  &  216   &  0.5278	&	0.5245	&	0.0369	&	\textBF{1.0000}	&	\textBF{1.0000}	&	1.0000\\
			parkinsons	&  22  &  147   &  0.1037	&	0.1109	&	0.0376	&	0.2778	&	\textBF{0.3889}	&	0.4509\\
			statlog-satimage	&  36  &  3594   &  0.5209	&	0.6722	&	0.9587	&	\textBF{0.9932}	&	0.9835	&	0.9914\\
			gisette	&  4971 &  3500  &  0.2544	&	0.0876	&	0.1066	&	0.8696	&	\textBF{0.9819}	&	0.9667\\
			waveform-1	& 21 &  3304 &  0.3039	&	\textBF{0.6456}	&	0.1564	&	0.2109	&	0.2061	&	0.2539\\
			madelon	&   500 & 1300  &  \textBF{1.0000}	&	0.1103	&	\textBF{1.0000}	&	0.9912	&	\textBF{1.0000}	&	1.0000\\
			synthetic-control-chart	& 60 &  400 &  0.7791	&	\textBF{1.0000}	&	0.3982	&	0.9778	&	0.9444	&	1.0000\\
			waveform-2	& 21 & 3304  &  0.2429	&	\textBF{0.5012}	&	0.3237	&	0.3607	&	0.3387	&	0.5368\\
			statlog-vehicle	& 18 &   629 &  0.4156	&	0.6411	&	0.7945	&	0.9363	&	\textBF{0.9618}	&	0.8729\\
			statlog-segment	& 18 &    1320 &  0.3764	&	\textBF{0.9049}	&	0.0700	&	0.3441	&	0.3413	&	0.3596\\
			\bottomrule
		\end{tabular}
	}
	\label{table:simulation_1_pr_0.7}
\end{table}

Our methods perform better compared to competition on majority of datasets. To compare the methods statistically, we use Friedman test based on average rank of methods; for  $\gamma \in \{0.9,0.5,0.3\}$. With $p$-val $<< 0.01$, we reject the null hypothesis that methods are equivalent across $\gamma$ values. We then proceed with Nemenyi post-hoc test to compare the algorithms pairwise and to find out the ones that differ significantly. The test identifies performance of two algorithms to be significantly different if their average ranks differ by at least the ``critical difference'' (CD). 

\begin{figure}[h!]
	\centering
	\vspace*{-.2in}
	\begin{subfigure}[b]{\linewidth}
		\centering
		\includegraphics[width=0.8\linewidth,height=3cm,]{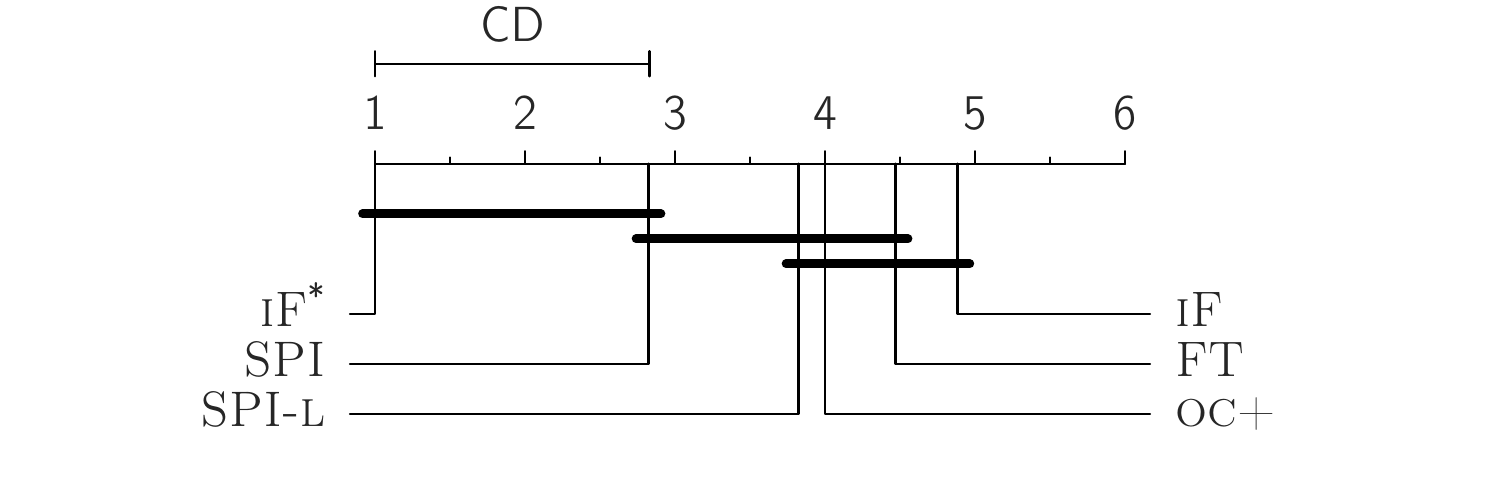}
		\caption{$\gamma = 0.9$}
	\end{subfigure}
	
	\begin{subfigure}[b]{\linewidth}
		\centering
		\includegraphics[width=0.8\linewidth,height=3cm,]{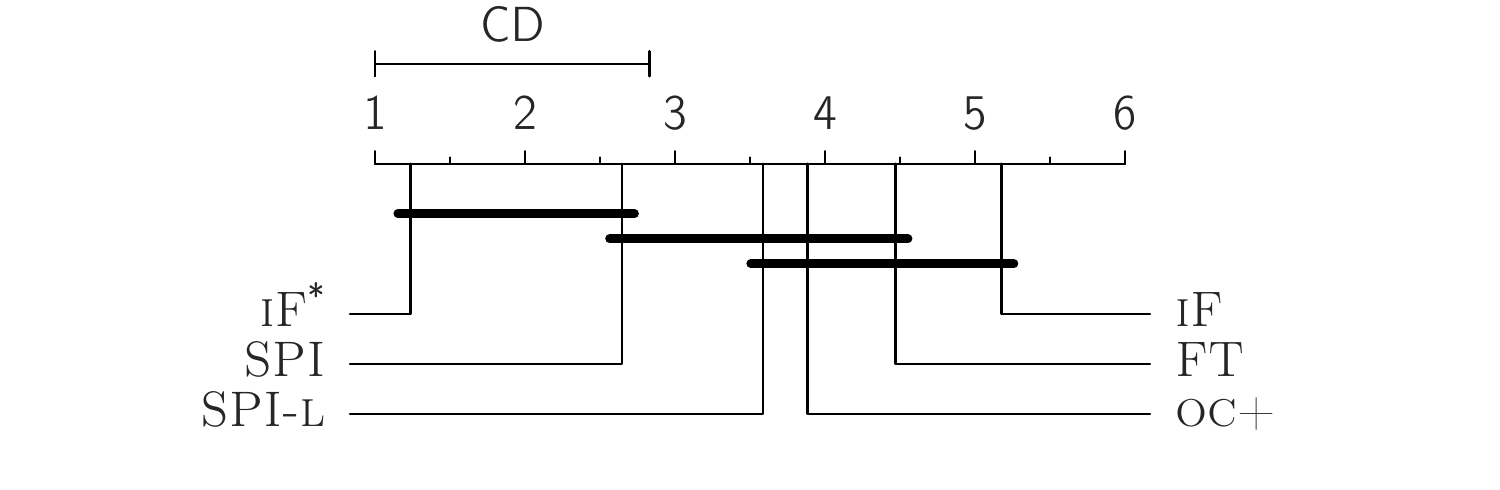}
		\caption{$\gamma = 0.5$}
	\end{subfigure}
	
	\begin{subfigure}[b]{\linewidth}
		\centering
		\includegraphics[width=0.8\linewidth,height=3cm,]{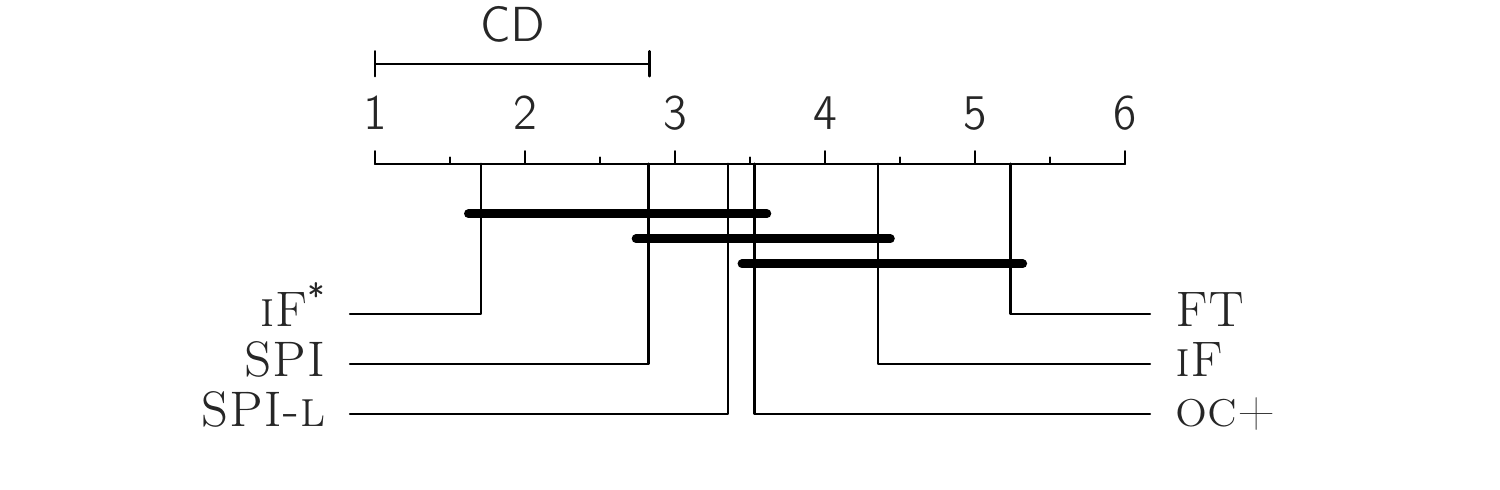}	
		\caption{$\gamma = 0.3$}
	\end{subfigure}
	\caption{Comparison of algorithms in terms of average rank~(wrt MAP) with the Nemenyi test on perturbed datasets with different $\gamma$ values. Groups of methods that are not significantly different (at $p$-val = $0.05$) are connected.}
	\label{fig:CD_diagrams_config1_all}
\end{figure}

Results of the post-hoc test are summarized through a graphical representation in Figure~\ref{fig:CD_diagrams_config1_all}. For $\gamma = 0.9$, we find that \iforestpi{} always ranks at $1$, since we retain 90\% of $X^*$ feature. Even with 10\% useful features in $X$ space, \spi{} is significantly better than \iforest{}, and  \spi{} has no significant difference when compared to \iforestpi{}. Other methods are comparable in performance to \iforest{} indicating that algorithms failed to exploit the minimal useful features present in $X$ space.

For $\gamma = 0.5$~(Figure~\ref{fig:CD_diagrams_config1_all}), we find that \spi{} has no significant difference when compared to \iforestpi{}. While all the PI incorporated methods are comparable, on average \spi{} and \spiscores{} rank better than the baselines.

For $\gamma=0.3$~(Figure~\ref{fig:CD_diagrams_config1_all}), we find that  \iforest{}~($X$-only) has no significant difference from \iforestpi{}~($X^*$ only). This is expected since $X$ space has many useful features.  Our methods have no significant difference from either \iforest{} or \iforestpi{}. While there is no significant difference, on average our methods still rank better than others.

\begin{figure}
	\centering
	\label{fig:metrics_evaluation_config_1_all}
	\begin{subfigure}[b]{0.24\linewidth}
		\centering
		\includegraphics[width=\linewidth]{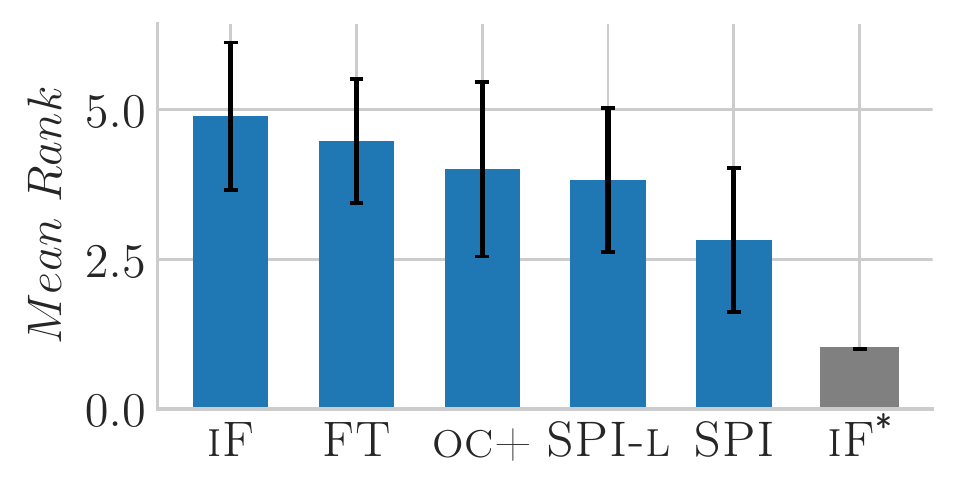}
	\end{subfigure}
	\begin{subfigure}[b]{0.24\linewidth}
		\centering
		\includegraphics[width=\linewidth]{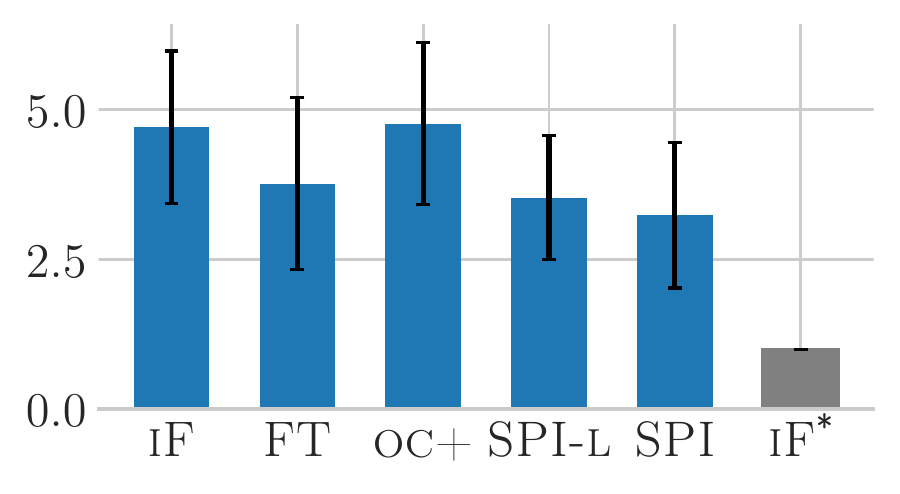}
	\end{subfigure}
	\begin{subfigure}[b]{0.24\linewidth}
		\centering
		\includegraphics[width=\linewidth]{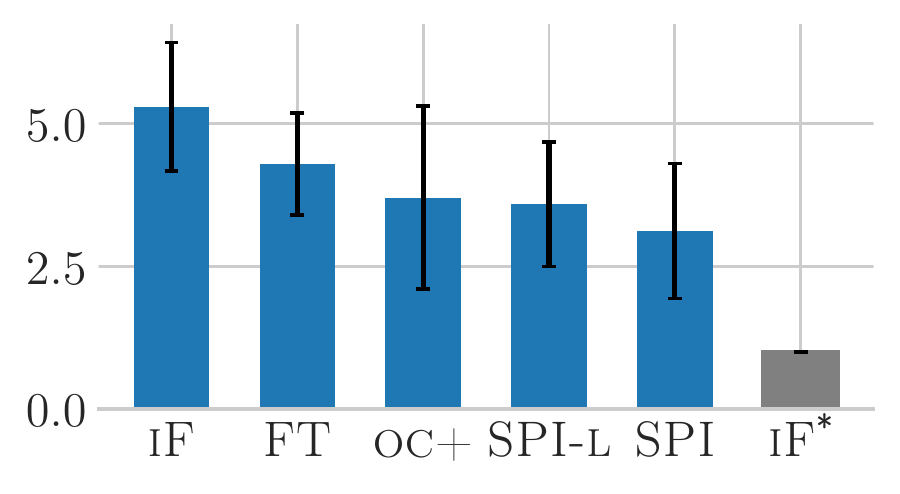}
	\end{subfigure}
	\begin{subfigure}[b]{0.24\linewidth}
		\centering
		\includegraphics[width=\linewidth]{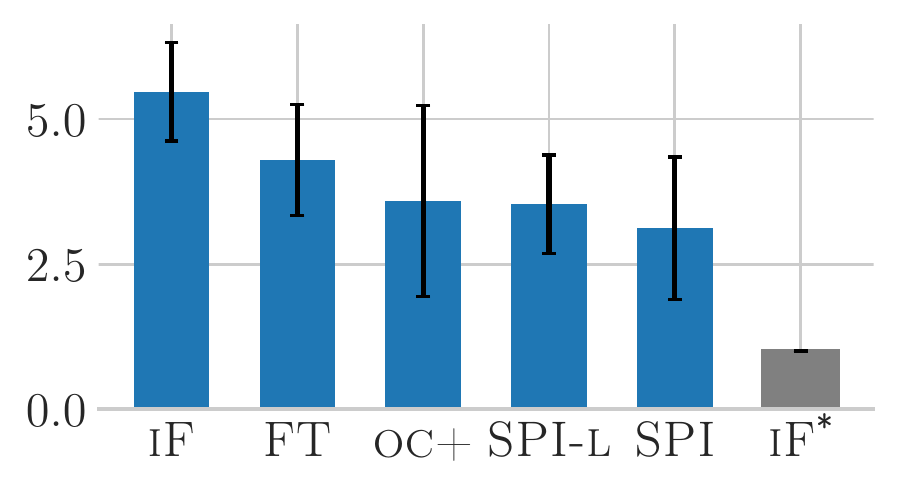}
	\end{subfigure}
	
	\begin{subfigure}[b]{0.24\linewidth}
		\centering
		\includegraphics[width=\linewidth]{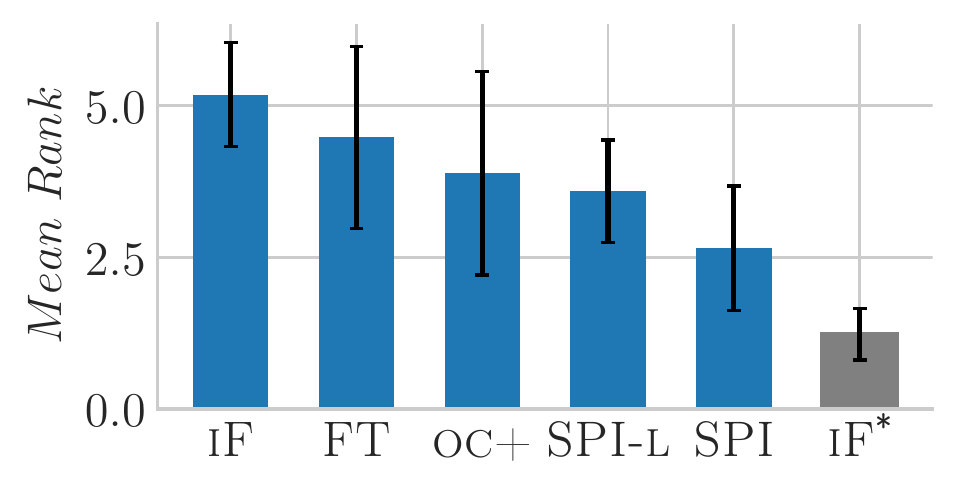}
	\end{subfigure}
	\begin{subfigure}[b]{0.24\linewidth}
		\centering
		\includegraphics[width=\linewidth]{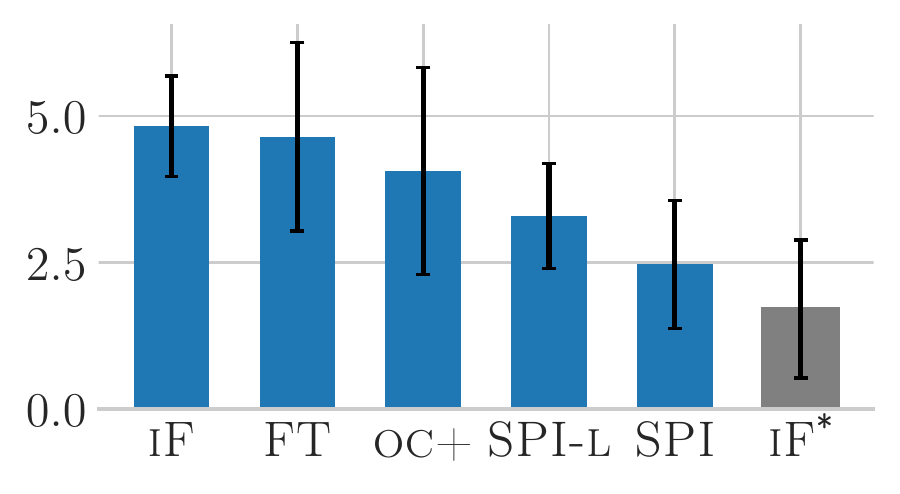}
	\end{subfigure}
	\begin{subfigure}[b]{0.24\linewidth}
		\centering
		\includegraphics[width=\linewidth]{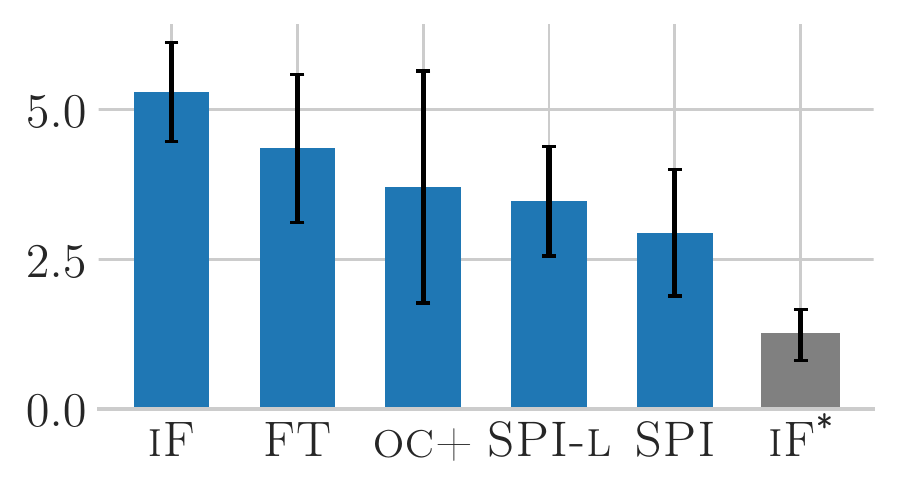}
	\end{subfigure}
	\begin{subfigure}[b]{0.24\linewidth}
		\centering
		\includegraphics[width=\linewidth]{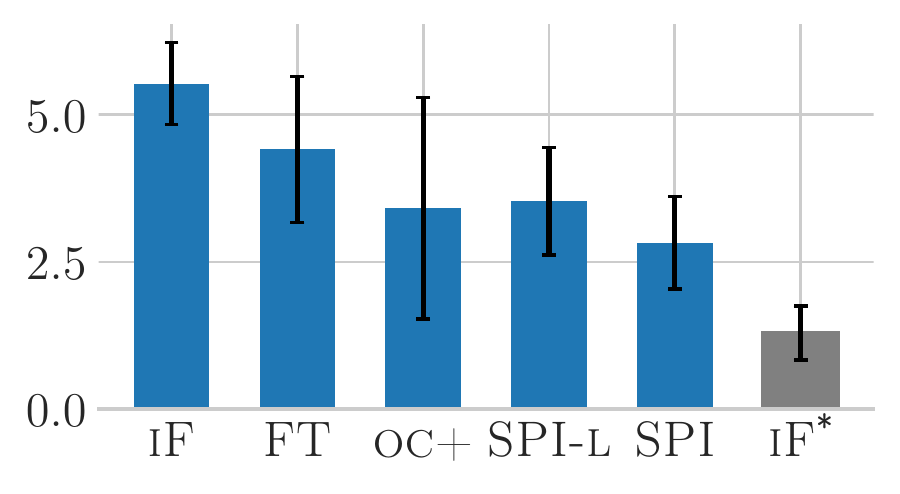}
	\end{subfigure}
	
	\begin{subfigure}[b]{0.24\linewidth}
		\centering
		\includegraphics[width=\linewidth]{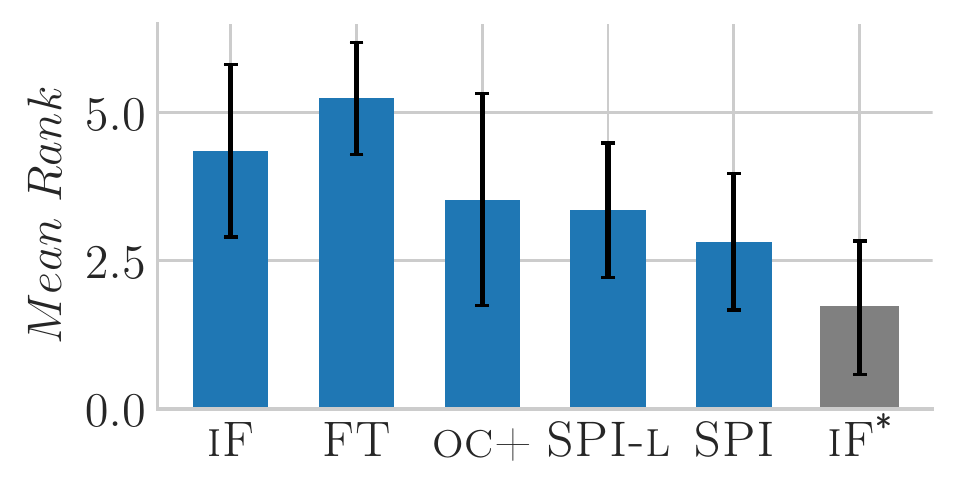}
		\caption{\textsc{MAP}}
	\end{subfigure}
	\begin{subfigure}[b]{0.24\linewidth}
		\centering
		\includegraphics[width=\linewidth]{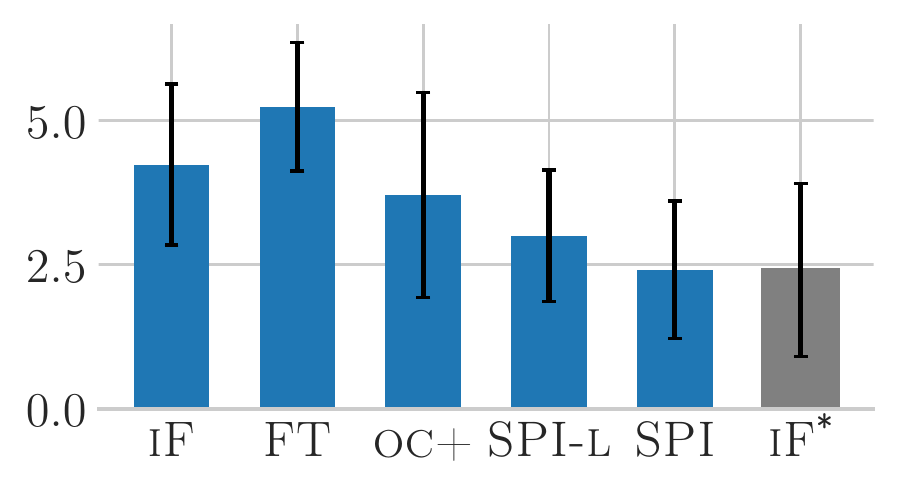}
		\caption{\textsc{AUC}}
	\end{subfigure}
	\begin{subfigure}[b]{0.24\linewidth}
		\centering
		\includegraphics[width=\linewidth]{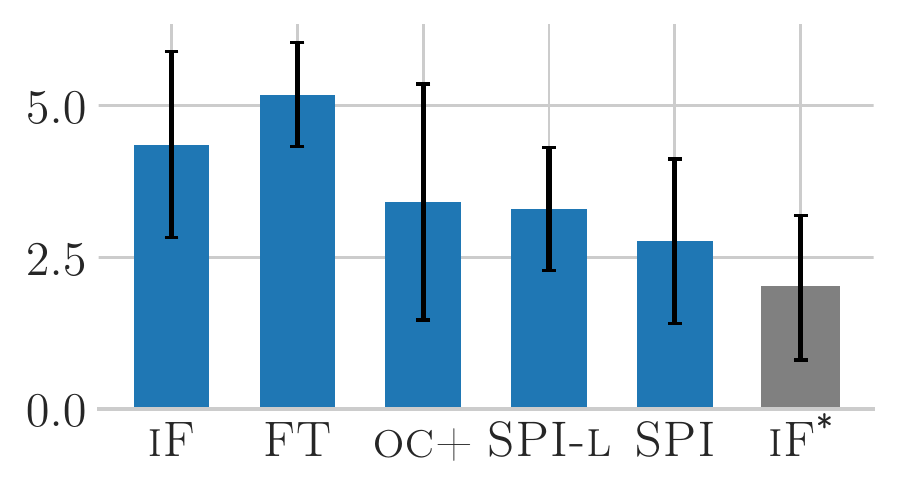}
		\caption{\textsc{ndcg@10}}
	\end{subfigure}
	\begin{subfigure}[b]{0.24\linewidth}
		\centering
		\includegraphics[width=\linewidth]{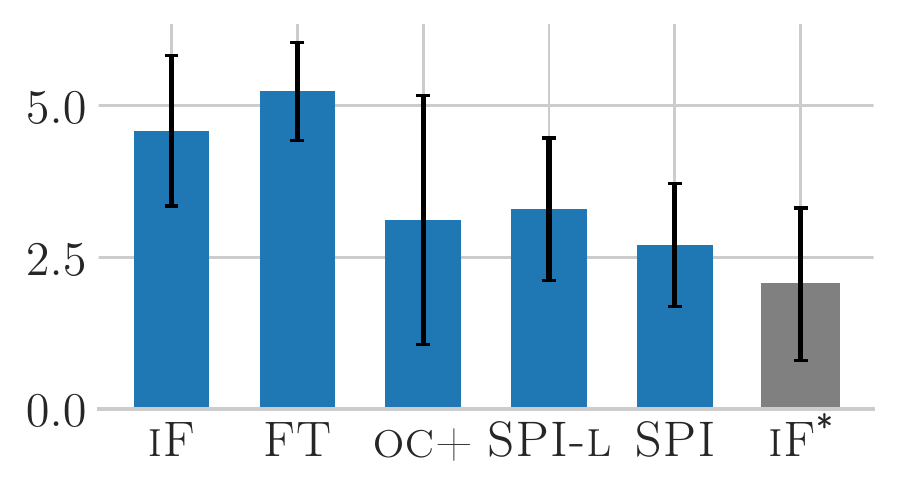}
		\caption{\textsc{precision@10}}
	\end{subfigure}
	\caption{\spi{} and \spiscores{} outperform competition w.r.t. different evaluation metrics row-wise for $\gamma \in \{0.9, 0.5, 0.3\} $. Average rank (bars) across benchmark datasets.  \iforestpi{} shown for reference.\label{fig:avg-ranks-all-config1}}
\end{figure}

We also report average rank of the algorithms against widely used ranking metrics including \textsc{AUC} of \textsc{ROC} curve, \textsc{ndcg@10} and \textsc{precision@10} in Figure~\ref{fig:avg-ranks-all-config1}. Notice that the results are consistent across measures for $\gamma \in \{0.9,0.5,0.3\}$, \spi{} and \spiscores{} performing among the best.

\subsection{Benchmark Evaluation: Simulation 2}
\label{subsec:appendix_simulation_2}

As mentioned earlier, the benchmark datasets do not have an explicit PI representation. Given a $p$ dimensional dataset, we identify two groups of features that represent $X$ space and $X^* $space in the following way.

$\boldsymbol{X}^*$ space: The original feature representation of the dataset is used as PI.

$\boldsymbol{X}$ space: We create $X$ space by generating new feature columns with Gaussian noise. In particular, we generate columns $10$ times of original dataset dimensionality with Gaussian noise $N(0.01\mu, 0.01\sigma)$ where $\mu$ is the mean and $\sigma$ is the standard deviation of all features in the original dataset.

Similar to first simulation setting, we construct $4$ versions per dataset with varying fraction $\gamma$ of original features~(PI) retained in $X^*$ space. In particular, each set has $\gamma p$ features in $X^*$, and $(1 - \gamma)p$+$10p$ features in $X$ for $\gamma \in \{0.9,0.7,0.5,0.3\}$.

\textit{Results}: We report the results on perturbed datasets with $\gamma \in \{0.9, 0.7,0.5,0.3\}$ as fraction of features retained in space $X^*$. Table~\ref{table:simulation_2_pr_0.1}, Table~\ref{table:simulation_2_pr}, Table~\ref{table:simulation_2_pr_0.5}, and Table~\ref{table:simulation_2_pr_0.7} report \textsc{MAP} against 7 datasets for different methods for $\gamma \in \{0.9, 0.7,0.5,0.3\}$. The results are averaged across 5 independent runs on stratified train-test splits.
\begin{table}
	\centering
	\caption{Mean Average Precision~(MAP) on perturbed datasets~(averaged over $5$ runs); for $\gamma=0.9$. \iforestpi{}~(reference) reports MAP in the $X^*$ space.}
	\resizebox{\columnwidth}{!}{%
		\begin{tabular}{lrr@{\hskip 0.1in}|@{\hskip 0.1in}rcrcr@{\hskip 0.1in} |@{\hskip 0.1in}r}
			\toprule
			Dataset& $p$ & Outliers &	\iforest & \ocsvm	&	\ft	&	\spiscores	&	\spi	&	\iforestpi\\
			\midrule
			breast-cancer	& 30  &  211   &	0.5159	&	0.2206	&	0.7139	&	0.7247	&	\textBF{0.7979}	&	0.7764\\
			ionosphere	& 33  &  125   &	0.3957	&	0.5208	&	0.3935	&	0.5377	&	\textBF{0.6939}	&	0.6984\\
			cardio	&  21 &  176   &	0.1240	&	0.1493	&	\textBF{0.5810}	&	0.3531	&	0.4355	&	0.6314\\
			satellite	& 36  &  2036   &	0.3771	&	0.3384	&	0.7233	&	\textBF{0.7388}	&	0.7347	&	0.7257\\
			wbc	&  30 &  21   &	0.1506	&	0.2530	&	0.2247	&	0.0995	&	\textBF{0.2994}	&	0.6208\\
			uci-ionosphere	& 33  &  126   &	0.3971	&	0.5462	&	0.4773	&	0.6034	&	\textBF{0.7128}	&	0.7703\\
			satimage-2	& 36  & 71   &	0.0206	&	0.0069	&	0.7425	&	0.6186	&	\textBF{0.6956}	&	0.9015\\
			\bottomrule
		\end{tabular}	
	}
	\label{table:simulation_2_pr_0.1}
\end{table}

\begin{table}
	\centering
	\caption{Mean Average Precision~(MAP) on perturbed datasets~(averaged over $5$ runs); for $\gamma=0.7$. \iforestpi{}~(reference) reports MAP in the $X^*$ space.}
	\resizebox{\columnwidth}{!}{%
		\begin{tabular}{lrr@{\hskip 0.1in}|@{\hskip 0.1in}rcrcr@{\hskip 0.1in} |@{\hskip 0.1in}r}
			\toprule
			Dataset&$p$ & Outliers &	\iforest & \ocsvm	&	\ft	&	\spiscores	&	\spi	&	\iforestpi\\
			\midrule
			breast-cancer	& 30  &  211   &	0.5714	&	0.2196	&	0.7324	&	0.6931	&	\textBF{0.7399}	&	0.7431\\
			ionosphere	& 33  &  125   &	0.4793	&	0.4545	&	0.3500	&	0.7041	&	\textBF{0.7043}	&	0.7134\\
			cardio	&  21 &  176   &	0.1962	&	0.0756	&	\textBF{0.8379}	&	0.5315	&	0.6518	&	0.8582\\
			satellite	& 36  &  2036   &	0.4446	&	0.2384	&	0.7056	&	\textBF{0.7217}	&	0.7060	&	0.7221\\
			wbc	&  30 &  21   &	0.1003	&	0.3702	&	\textBF{0.6000}	&	0.2624	&	0.5563	&	0.6204\\
			uci-ionosphere	& 33  &  126   &	0.4253	&	0.3946	&	0.4391	&	0.6609	&	\textBF{0.7353}	&	0.8214\\
			satimage-2	& 36  & 71   &	0.0871	&	0.0042	&	0.7377	&	0.8017	&	\textBF{0.8400}	&	0.9481\\
			\bottomrule
		\end{tabular}	
	}
	\label{table:simulation_2_pr}
\end{table}

\begin{table}
	\centering
	\caption{Mean Average Precision~(MAP) on perturbed datasets~(averaged over $5$ runs); for $\gamma=0.5$. \iforestpi{}~(reference) reports MAP in the $X^*$ space.}
	\resizebox{\columnwidth}{!}{%
		\begin{tabular}{lrr@{\hskip 0.1in}|@{\hskip 0.1in}rcrcr@{\hskip 0.1in} |@{\hskip 0.1in}r}
			\toprule
			Dataset& $p$ & Outliers &	\iforest & \ocsvm	&	\ft	&	\spiscores	&	\spi	&	\iforestpi\\
			\midrule
			breast-cancer	& 30  &  211   &	0.5740	&	0.2196	&	\textBF{0.7981}	&	0.7224	&	0.7214	&	0.7656\\
			ionosphere	& 33  &  125   &	0.5644	&	0.3429	&	0.3378	&	0.6246	&	\textBF{0.6933}	&	0.7213\\
			cardio	&  21 &  176   &	0.2209	&	0.0950	&	\textBF{0.6250}	&	0.5380	&	0.6108	&	0.6082\\
			satellite	& 36  &  2036   &	0.5627	&	0.2359	&	0.6718	&	\textBF{0.7052}	&	0.6660	&	0.7292\\
			wbc	&  30 &  21   &	0.1804	&	0.2322	&	\textBF{0.5829}	&	0.3817	&	0.5103	&	0.6094\\
			uci-ionosphere	& 33  &  126   &	0.5172	&	0.3871	&	0.4339	&	0.6240	&	\textBF{0.7212}	&	0.8234\\
			satimage-2	& 36  & 71   &	0.1675	&	0.0042	&	0.7353	&	0.8101	&	\textBF{0.8029}	&	0.8882\\
			\bottomrule
		\end{tabular}	
	}
	\label{table:simulation_2_pr_0.5}
\end{table}

\begin{table}
	\centering
	\caption{Mean Average Precision~(MAP) on perturbed datasets~(averaged over $5$ runs); for $\gamma=0.3$. \iforestpi{}~(reference) reports MAP in the $X^*$ space.}
	\resizebox{\columnwidth}{!}{%
		\begin{tabular}{lrr@{\hskip 0.1in}|@{\hskip 0.1in}rcrcr@{\hskip 0.1in} |@{\hskip 0.1in}r}
			\toprule
			Dataset& $p$ & Outliers &	\iforest & \ocsvm	&	\ft	&	\spiscores	&	\spi	&	\iforestpi\\
			\midrule
			breast-cancer	& 30  &  211   &	0.6108	&	0.2193	&	\textBF{0.6706}	&	0.6682	&	0.6272	&	0.6122\\
			ionosphere	& 33  &  125   &	0.6045	&	0.2817	&	0.3587	&	0.6683	&	\textBF{0.7198}	&	0.7230\\
			cardio	&  21 &  176   &	0.3305	&	0.1171	&	\textBF{0.5851}	&	0.5044	&	0.5230	&	0.5254\\
			satellite	& 36  &  2036   &	0.6029	&	0.2359	&	0.6662	&	\textBF{0.7034}	&	0.6530	&	0.6726\\
			wbc	&  30 &  21   &	0.2769	&	0.2115	&	0.4791	&	\textBF{0.4823}	&	0.4419	&	0.4759\\
			uci-ionosphere	& 33  &  126   &	0.6244	&	0.3728	&	0.4420	&	0.6830	&	\textBF{0.7225}	&	0.7768\\
			satimage-2	& 36  & 71   &	0.3827	&	0.0043	&	0.7450	&	0.7949	&	\textBF{0.8305}	&	0.8327\\
			\bottomrule
		\end{tabular}	
	}
	\label{table:simulation_2_pr_0.7}
\end{table}

To compare the methods statistically, we use Friedman test based on average rank of methods; for  $\gamma \in \{0.9, 0.7,0.5,0.3\}$. With $p$-val $<< 0.01$, we reject the null hypothesis that methods are equivalent across $\gamma$ values. We proceed with the Nemenyi post hoc test to find out which methods actually differ.

\begin{figure}
	\centering
	\begin{subfigure}[b]{\linewidth}
		\centering
		\includegraphics[width=0.8\linewidth,height=3cm,]{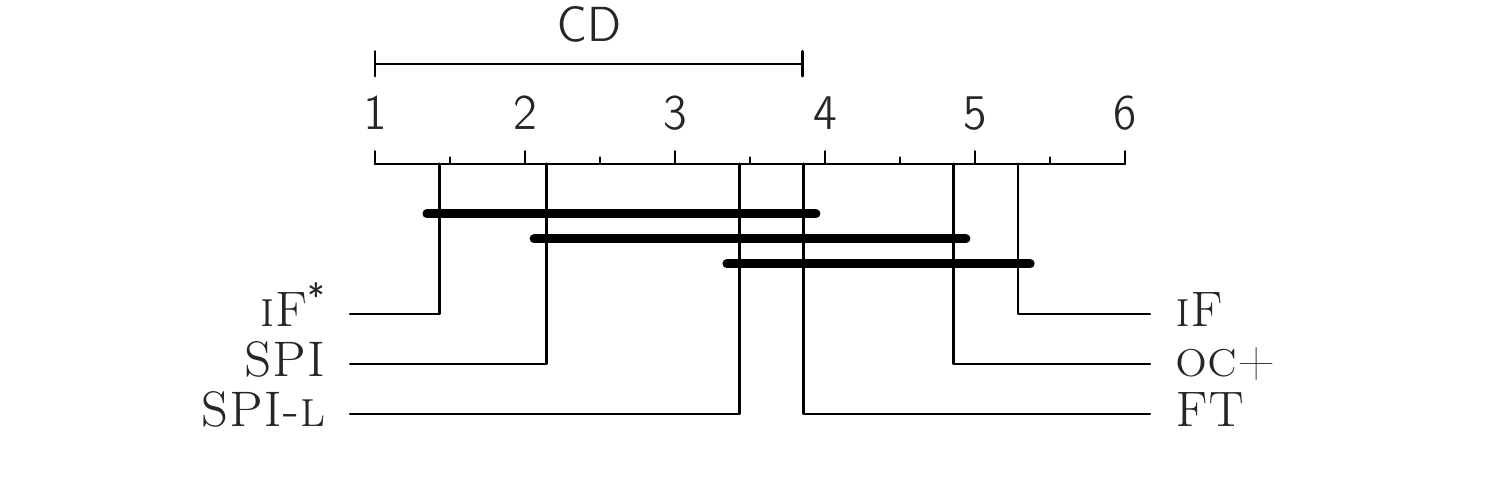}
		\caption{$\gamma = 0.9$} 
	\end{subfigure}
	\vspace*{0.1in}
	\begin{subfigure}[b]{\linewidth}
		\centering
		\includegraphics[width=0.8\linewidth,height=3cm,]{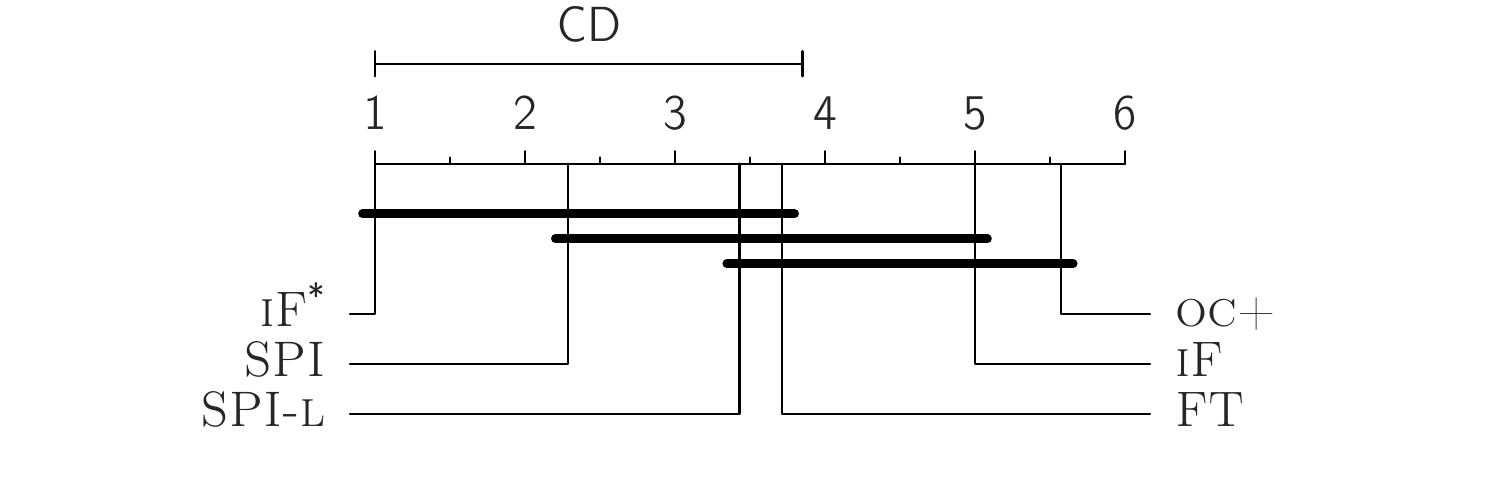}	
		\caption{$\gamma = 0.7$} 
	\end{subfigure}	
	\vspace*{0.1in}
	\begin{subfigure}[b]{\linewidth}
		\centering
		\includegraphics[width=0.8\linewidth,height=3cm,]{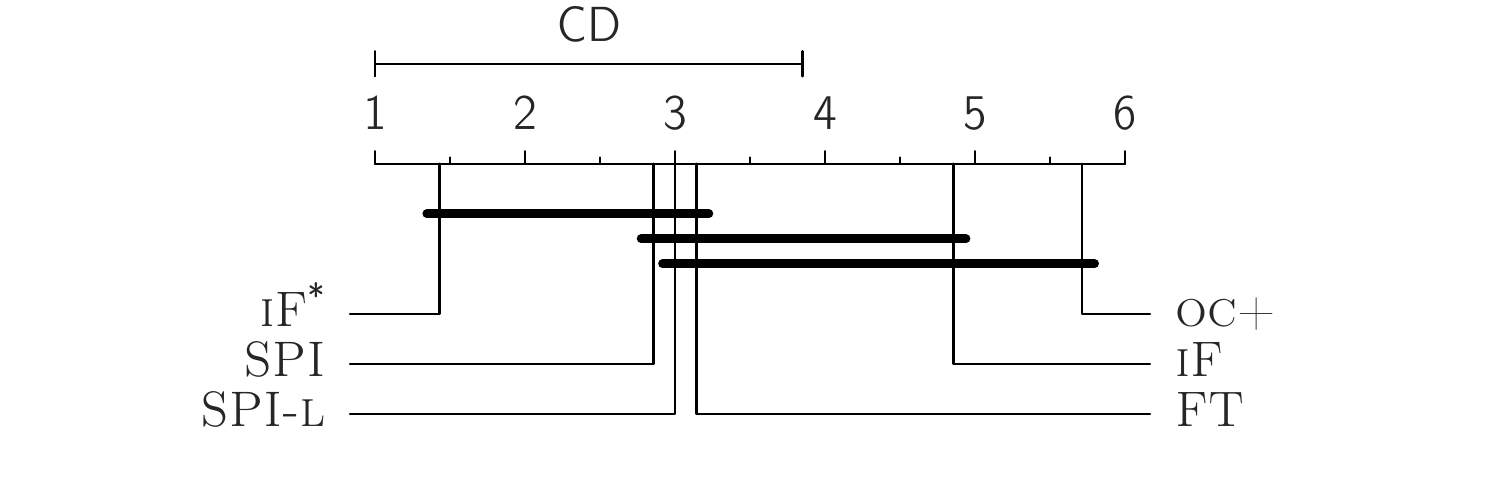}
		\caption{$\gamma = 0.5$}
	\end{subfigure}
	\vspace*{0.1in}
	\begin{subfigure}[b]{\linewidth}
		\centering
		\includegraphics[width=0.8\linewidth,height=3cm,]{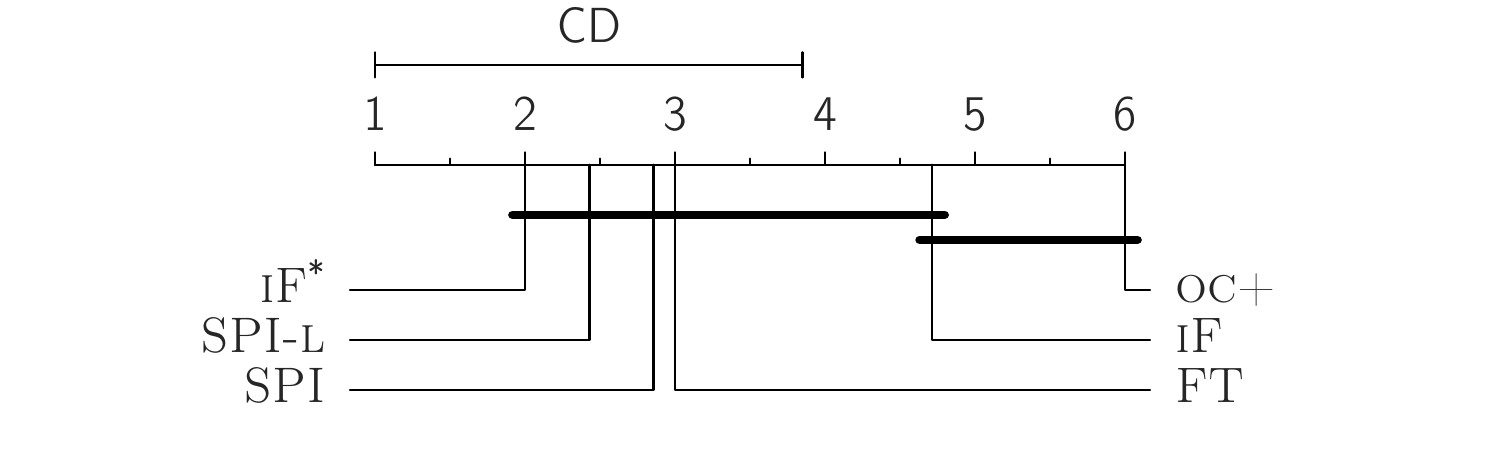}	
		\caption{$\gamma = 0.3$} 
	\end{subfigure}
	\caption{Simulation $2$: Comparison of algorithms in terms of average rank~(wrt MAP) with the Nemenyi test for different $\gamma$ values. Groups of methods that are not significantly different (at $p$-val = $0.05$) are connected.\label{fig:CD_diagrams_config2_all}}
\end{figure}

We summarize our finding after post hoc test graphically in Figure~\ref{fig:CD_diagrams_config2_all}. For $\gamma = 0.9$, we find that all the PI incorporated methods except \oc{} have no significant difference when compared to \iforestpi{}. Our method \spi{} significantly outperforms \iforest{}. While the PI incorporated methods are comparable, on average our methods rank better than others.

For $\gamma = 0.7$ and $\gamma = 0.5$~(Figure~\ref{fig:CD_diagrams_config2_all}), we find that methods incorporating PI through knowledge transfer have no significant difference to \iforestpi{}.  While all the PI incorporated methods are comparable except \oc{}, comparatively on average our methods rank better.

For $\gamma=0.3$~(Figure~\ref{fig:CD_diagrams_config2_all}), we find that  \iforest{}~($X$-only) has no significant difference from \iforestpi{}~($X^*$ only). This is expected since $X$ space has many useful features.  Our methods have no significant difference from either \iforest{} or \iforestpi{}. On average our methods still rank better than others.

\begin{figure}
	\centering
	\begin{subfigure}[b]{0.24\linewidth}
		\centering
		\includegraphics[width=\linewidth]{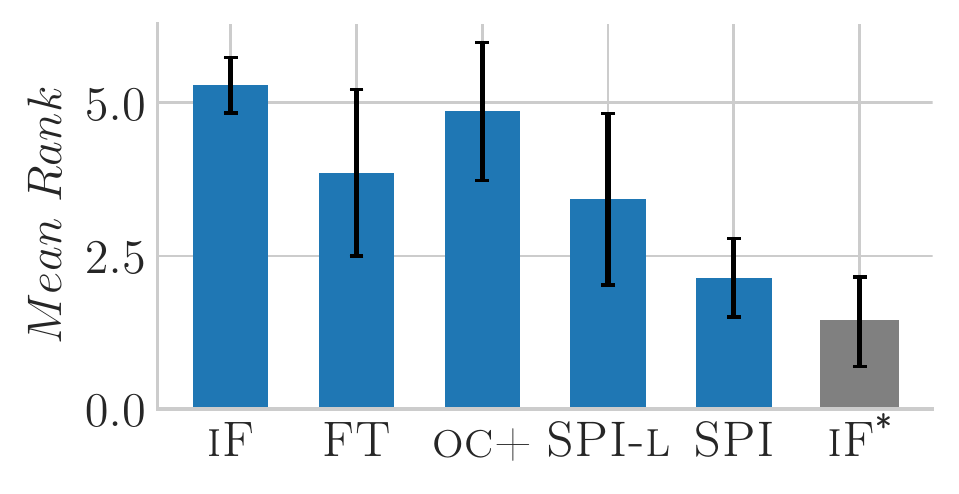}
	\end{subfigure}
	\begin{subfigure}[b]{0.24\linewidth}
		\centering
		\includegraphics[width=\linewidth]{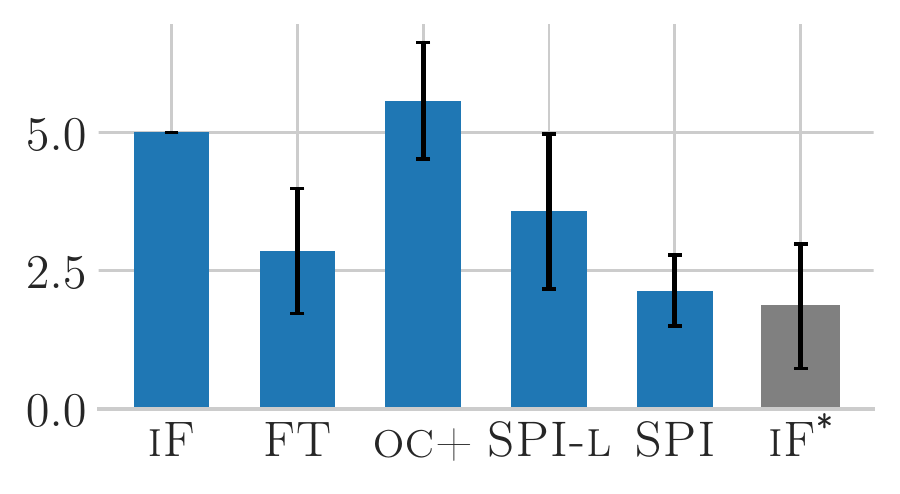}
	\end{subfigure}
	\begin{subfigure}[b]{0.24\linewidth}
		\centering
		\includegraphics[width=\linewidth]{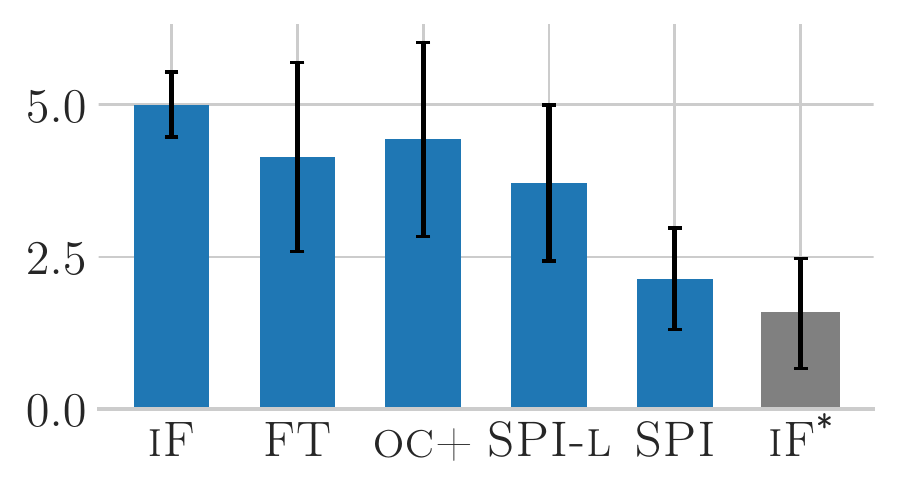}
	\end{subfigure}
	\begin{subfigure}[b]{0.24\linewidth}
		\centering
		\includegraphics[width=\linewidth]{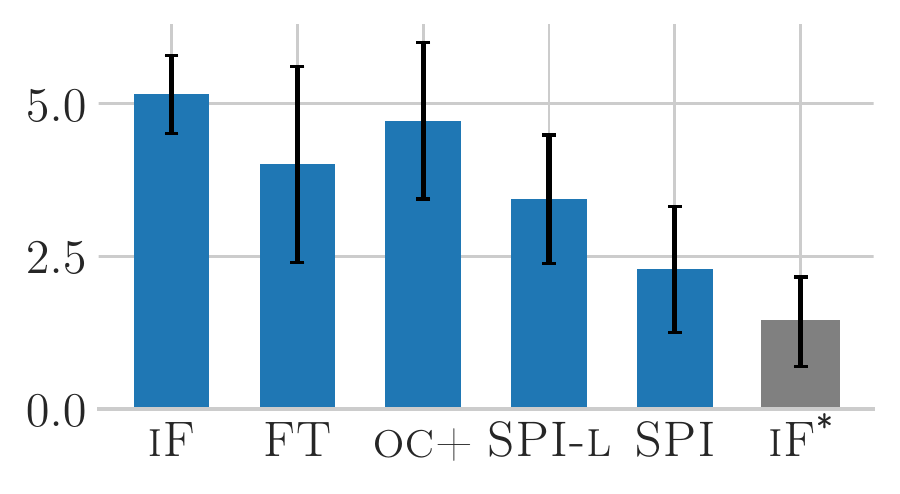}
	\end{subfigure}
	
	\begin{subfigure}[b]{0.24\linewidth}
		\centering
		\includegraphics[width=\linewidth]{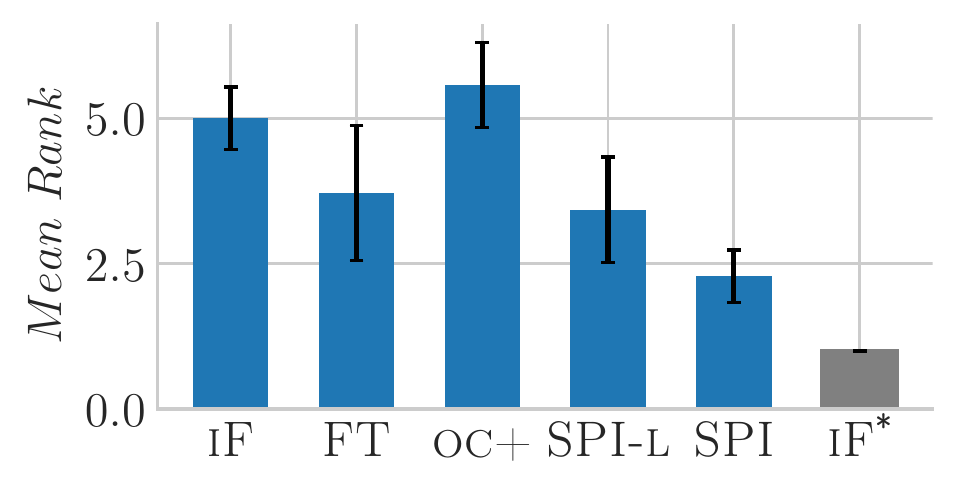}
	\end{subfigure}
	\begin{subfigure}[b]{0.24\linewidth}
		\centering
		\includegraphics[width=\linewidth]{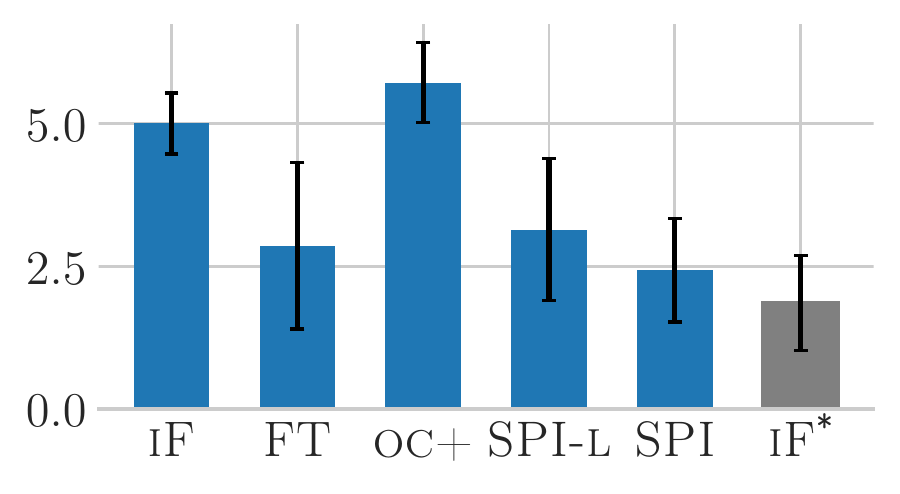}
	\end{subfigure}
	\begin{subfigure}[b]{0.24\linewidth}
		\centering
		\includegraphics[width=\linewidth]{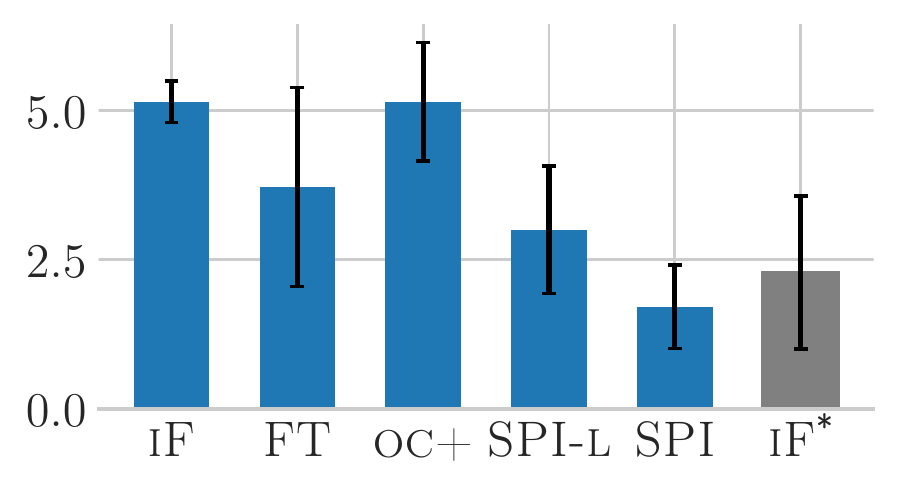}
	\end{subfigure}
	\begin{subfigure}[b]{0.24\linewidth}
		\centering
		\includegraphics[width=\linewidth]{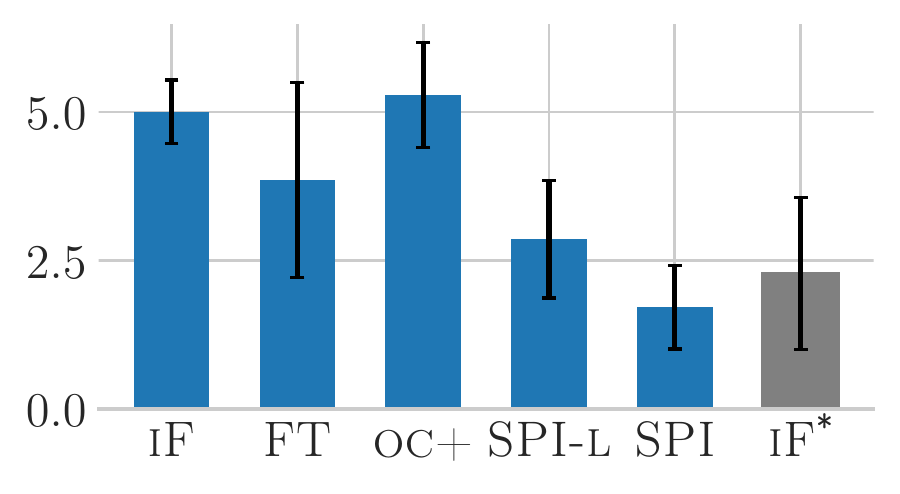}
	\end{subfigure}
	
	\begin{subfigure}[b]{0.24\linewidth}
		\centering
		\includegraphics[width=\linewidth]{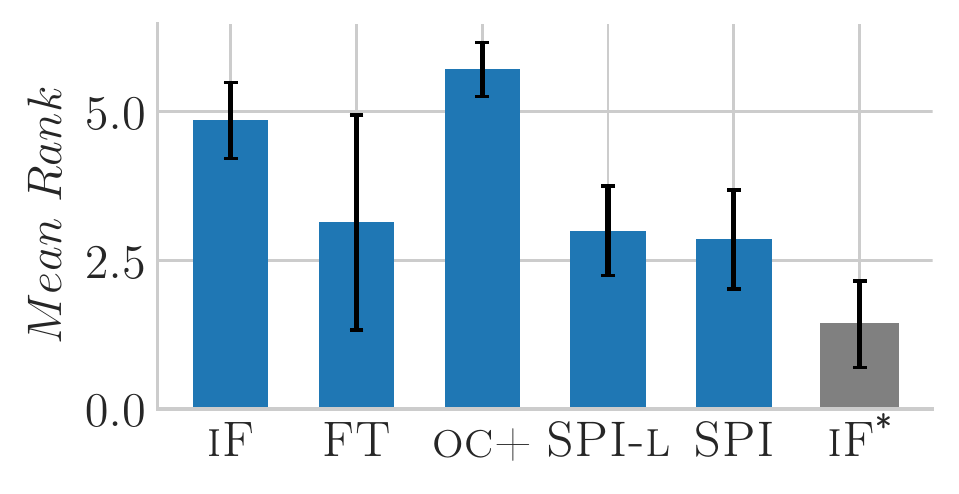}
	\end{subfigure}
	\begin{subfigure}[b]{0.24\linewidth}
		\centering
		\includegraphics[width=\linewidth]{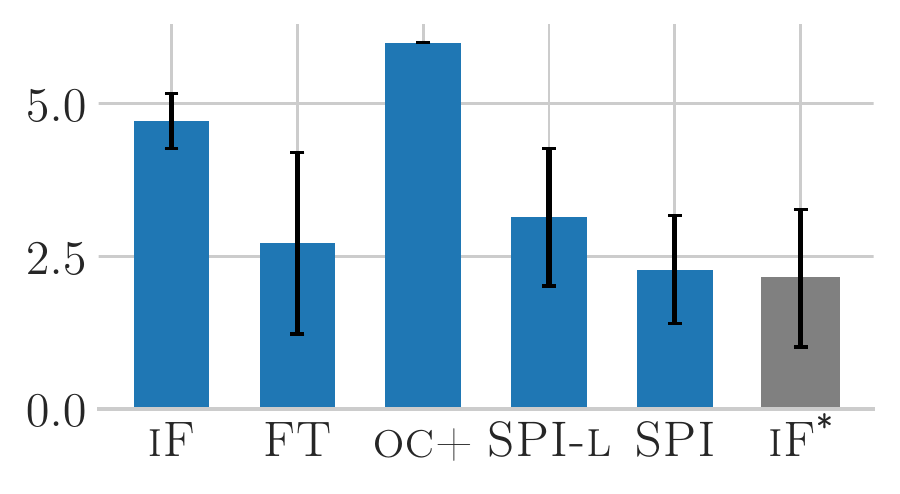}
	\end{subfigure}
	\begin{subfigure}[b]{0.24\linewidth}
		\centering
		\includegraphics[width=\linewidth]{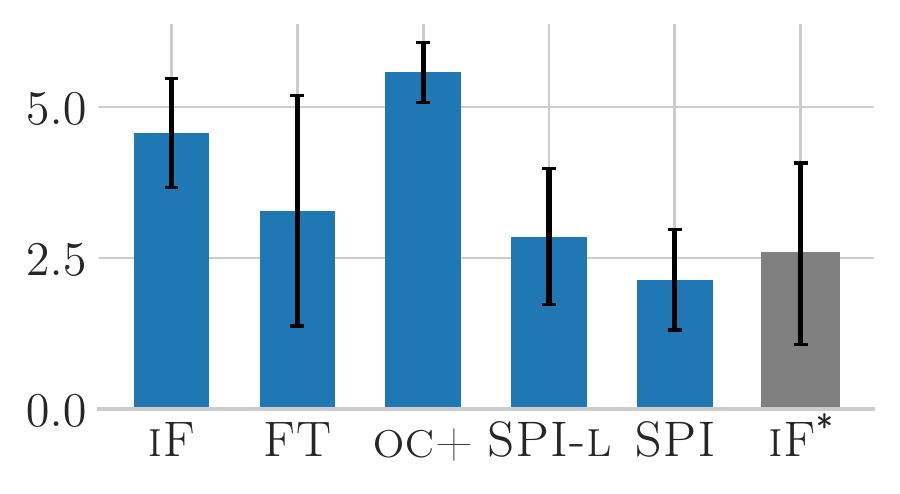}
	\end{subfigure}
	\begin{subfigure}[b]{0.24\linewidth}
		\centering
		\includegraphics[width=\linewidth]{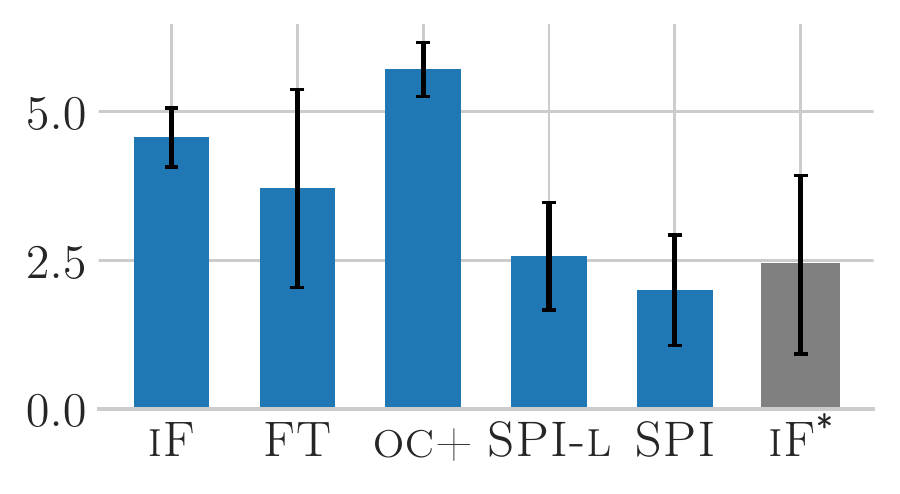}
	\end{subfigure}
	
	\begin{subfigure}[b]{0.24\linewidth}
		\centering
		\includegraphics[width=\linewidth]{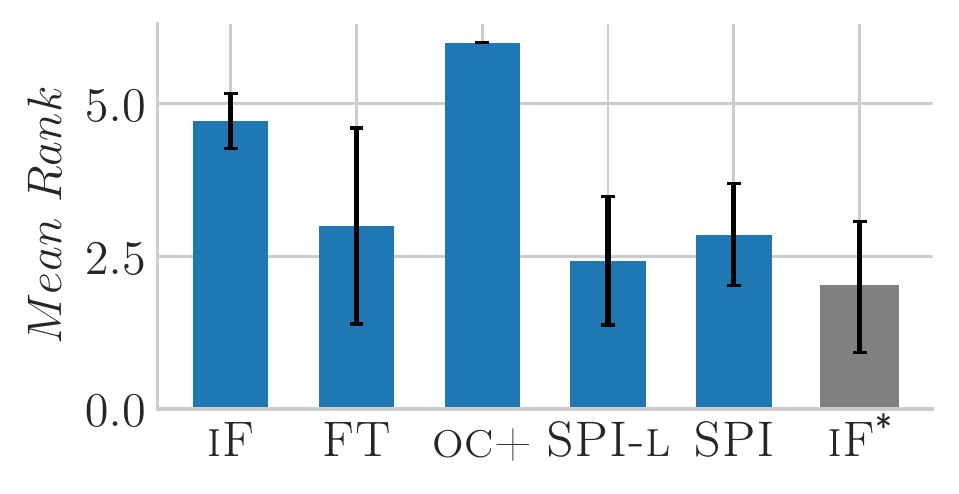}
		\caption{\textsc{MAP}}
	\end{subfigure}
	\begin{subfigure}[b]{0.24\linewidth}
		\centering
		\includegraphics[width=\linewidth]{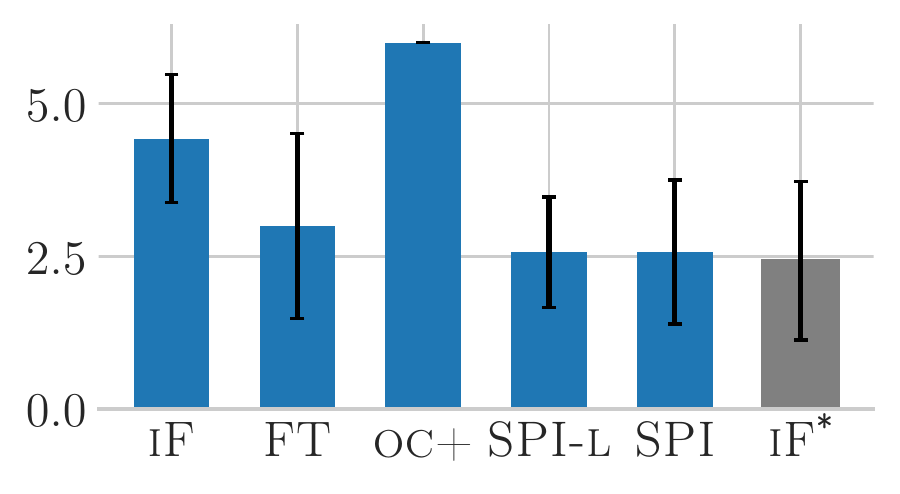}
		\subcaption{\textsc{AUC}}
	\end{subfigure}
	\begin{subfigure}[b]{0.24\linewidth}
		\centering
		\includegraphics[width=\linewidth]{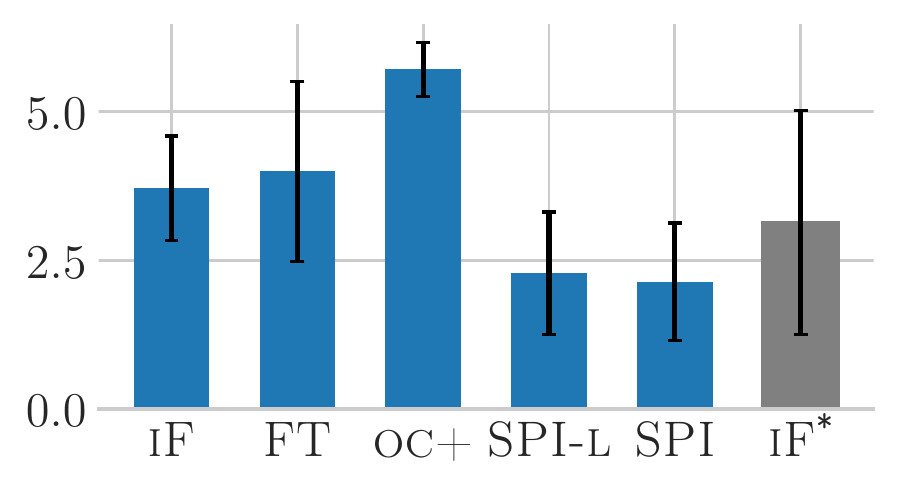}
		\caption{\textsc{ndcg@10}}
	\end{subfigure}
	\begin{subfigure}[b]{0.24\linewidth}
		\centering
		\includegraphics[width=\linewidth]{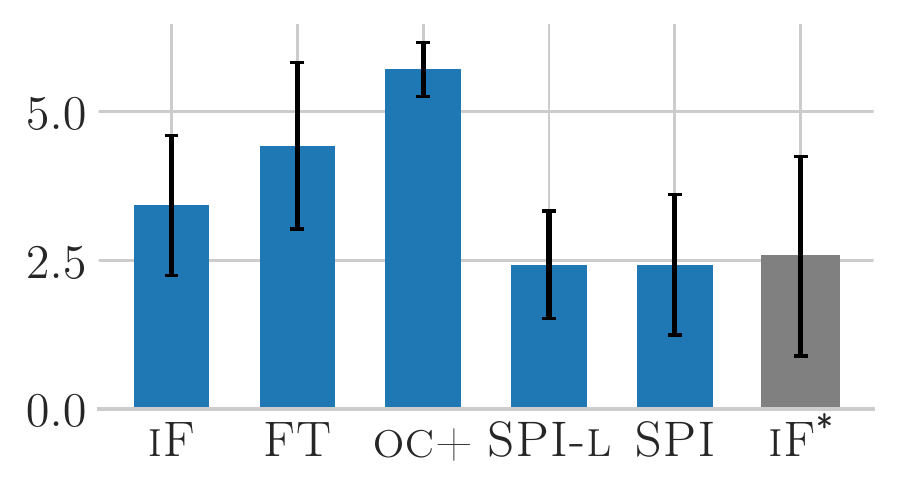}
		\caption{\textsc{precision@10}}
	\end{subfigure}
	\caption{Simulation 2: 
		\spi{} and \spiscores{} rank better than competing methods w.r.t. different evaluation metrics row-wise for $\gamma \in \{0.9, 0.7, 0.5, 0.3\} $. Average rank (bars) across benchmark datasets. \iforestpi{} shown for reference.\label{fig:metrics_evaluation_config_2_all}}
\end{figure}

We also report average rank of the algorithms against other popular ranking metrics including \textsc{AUC} of \textsc{ROC} curve, \textsc{ndcg@10} and \textsc{precision@10} in Figure~\ref{fig:metrics_evaluation_config_2_all}. Notice that the results are consistent across measures for $\gamma \in \{0.9,0.7, 0.5, $ $0.3\}$, \spi{} and \spiscores{} rank better than competing methods.
\end{document}